\documentclass[conference]{IEEEtran}
\usepackage{amsmath,amssymb,amsfonts}
\usepackage{graphicx}
\usepackage{textcomp}
\usepackage{xcolor}
\def\BibTeX{{\rm B\kern-.05em{\sc i\kern-.025em b}\kern-.08em
    T\kern-.1667em\lower.7ex\hbox{E}\kern-.125emX}}

\usepackage{biblatex}
\usepackage{mathtools, amssymb, amsthm}
\usepackage{pgfplots}
\usepackage{multirow}
\usepackage{booktabs}
\usepackage{graphicx}
\usepackage{pgfplotstable}
\usepackage{csvsimple}
\usepackage{hhline}
\usepackage[printonlyused,withpage]{acronym}
\usepackage{tabularx}
\usepackage[figuresright]{rotating}
\usepackage{makecell}
\usepackage{siunitx,etoolbox}
\usepackage[nomessages]{fp}
\pgfplotsset{compat=1.18}
\usepgfplotslibrary{fillbetween}
\usepackage{caption}
\usepackage{subcaption}
\usepgfplotslibrary{groupplots}
\usepackage{amsmath}
\usepackage{rotating}
\usepackage{placeins}
\usepackage[export]{adjustbox}
\usepackage{algpseudocode}
\usepackage[linesnumbered]{algorithm2e}
\usepackage{placeins}
\usepackage{xcolor}
\usepackage[dvipsnames]{xcolor}
\usepackage{afterpage}
\usepackage{amsmath}
\usepackage{hyperref}
\usepackage{dsfont}

\usepackage{makecell}

\colorlet{false}{red!100}
\colorlet{correct}{Green!100}

\definecolor{colorgptv}{named}{blue}
\definecolor{colorgpto}{named}{green}
\definecolor{colorpremise}{named}{gray}
\definecolor{colorcadprompt}{named}{red}

\renewcommand{\IEEEauthorrefmark}[1]{%
  \textsuperscript{\normalfont\footnotesize
    \ifnum#1=1 1\fi
    \ifnum#1=2 2\fi
    \ifx#1\string1,2 1,2\fi
  }%
}

\begin{document}

\title{EvoCAD: Evolutionary CAD Code Generation with Vision Language Models}

\author{\IEEEauthorblockN{Tobias Preintner\IEEEauthorrefmark{1}\textsuperscript{,}\IEEEauthorrefmark{2},
Weixuan Yuan\IEEEauthorrefmark{2},
Adrian König\IEEEauthorrefmark{2}, 
Thomas Bäck\IEEEauthorrefmark{1},
Elena Raponi\IEEEauthorrefmark{1},
Niki van Stein\IEEEauthorrefmark{1}}
\IEEEauthorblockA{\IEEEauthorrefmark{1}Institute of Advanced Computer Science, Leiden University, Leiden, The Netherlands}
\IEEEauthorblockA{\IEEEauthorrefmark{2}BMW Group, Munich, Germany}}%

\maketitle

\begin{abstract}
Combining large language models with evolutionary computation algorithms represents a promising research direction leveraging the remarkable generative and in-context learning capabilities of LLMs with the strengths of evolutionary algorithms.
In this work, we present EvoCAD, a method for generating computer-aided design (CAD) objects through their symbolic representations using vision language models and evolutionary optimization. 
Our method samples multiple CAD objects, which are then optimized using an evolutionary approach with vision language and reasoning language models.
We assess our method using GPT-4V and GPT-4o, evaluating it on the CADPrompt benchmark dataset and comparing it to prior methods.
Additionally, we introduce two new metrics based on topological properties defined by the Euler characteristic, which capture a form of semantic similarity between 3D objects.
Our results demonstrate that EvoCAD outperforms previous approaches on multiple metrics, particularly in generating topologically correct objects, which can be efficiently evaluated using our two novel metrics that complement existing spatial metrics.



%
\end{abstract}


\section{Introduction}
%
The use of generative AI tools powered by large language models (LLMs) has transformed the way humans work, create, and develop.
%
%
However, while significant attention is directed towards textual knowledge tasks, comparatively little focus is devoted on working with symbolic representations, such as those utilized in computer-aided design (CAD).
These code-like textual representations, in the following referred as CAD code, enable visual assets to be processed by LLMs \cite{qiu2025canllm}. This creates opportunities to enhance productivity and efficiency in design and manufacturing.

While the final outcome is intended to be a 3D object, generating CAD code instead of directly creating a 3D object, such as point clouds or meshes, has various advantages.
CAD code can create fully parametric models using a very small amount of code, thereby requiring less memory than most other 3D representations, while also being able to output high-quality CAD formats such as STEP and AMF, in addition to traditional STL \cite{cadquery2024}.
%
Specifically, generating CAD code from a user prompt is well-suited for leveraging the powerful text-to-text generation capabilities of LLMs.

However, unlike pure code generation for coding tasks, generating 3D objects through CAD code requires additional visual evaluation.
This can be done manually by compiling the generated CAD code into a 3D object, evaluating the object, and providing feedback to the LLM for further improvement.
%
Recent works \cite{yuan2023-3dpremise, alrashedy2025cadprompt} have automated this manual feedback process by utilizing vision language models (VLMs).
%
However, their methods primarily aim to remove humans from the feedback loop. In this paper, we propose and investigate a novel approach to enhance CAD code generation by combining the strengths of evolutionary computation and VLMs.

%
Our method samples multiple CAD codes, which are subsequently optimized using an evolutionary approach.
To achieve this, we use a technique for ranking text-to-3D alignment within a population of CAD objects using VLMs and reasoning language models (RLMs). 
Followed by LLM-based crossover and mutation operations, we can leverage evolutionary optimization demonstrating superior performance on the CADPrompt \cite{alrashedy2025cadprompt} benchmark dataset.
To the best of our knowledge, we are the first to present and investigate evolutionary CAD code generation with VLMs. 

Additionally, we propose a novel type of metric for comparing 3D objects using topological properties, which is particularly useful for CAD objects that often contain complex structures and diverse topological characteristics.
In this way, our topology-based metrics assess a form of semantic similarity between two objects, complementing existing metrics that only evaluate spatial properties.
Our code is available at \mbox{\textit{https://github.com/toprei/evo-cad}}.

Our contributions include:
\begin{itemize}
\item A novel method that combines LLMs and evolutionary computation to generate CAD objects through code.
\item A new type of topology-based metric for 3D object evaluation, complementing existing geometrical metrics.
\item Evaluation using GPT-4V and GPT-4o on the CADPrompt benchmark, showing cross-model applicability.
\item A comparison of EvoCAD with prior methods, demonstrating the superiority of our approach.
\item A technique for ranking text-to-3D alignment using vision language and reasoning language models.
\end{itemize}

\label{chapter:introduction}

\section{Related Work}
\begin{figure*}[htbp]
\centering
    \includegraphics[width=\linewidth]{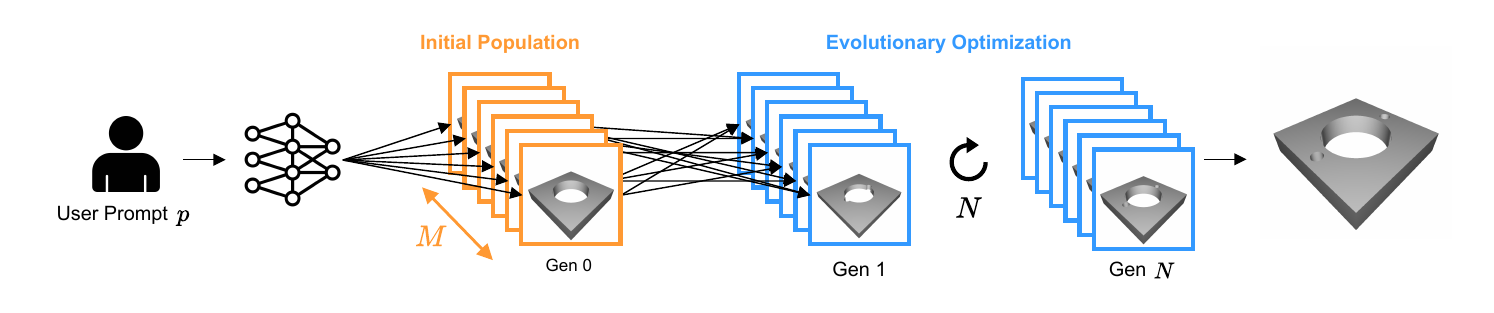}%
    \caption{Visual illustration of our evolutionary CAD code generation method. Beginning with a textual user prompt describing the desired object, our approach initializes a population of CAD codes using a large language model, followed by evolutionary optimization with vision and reasoning language models. \label{fig:overview}}
\end{figure*}




\subsection{Evolutionary Algorithms with Large Language Models}
Evolutionary algorithms are founded upon the principle of evolution \cite{michalewicz2013genetic}. 
A promising research directions integrates the remarkable generation and in-context learning abilities of LLMs with evolutionary computation using LLM-based crossover and mutation operators \cite{chauhan2025evolutionary}.

Eureka \cite{ma2024eureka} generates reinforcement learning reward functions that outperform expert human-engineered rewards.
FunSearch \cite{romera2024funsearch} was the first evolutionary loop to evolve small heuristics, EoH \cite{liu2024eoh} made this process more efficient, and later research efforts such as LLaMEA \cite{van2024llamea}, LLaMEA-HPO \cite{van2024loop} and AlphaEvolve \cite{novikov2025alphaevolve} are generating and evolving complete algorithms and solutions, spanning many lines of code.
%
%
In neural architecture search, \cite{morris2024llmnas} autonomously produced model variants with improved accuracy using an LLM-guided evolution.
%
GAVEL \cite{todd2024gavel} generates novel games by mutating and recombining games that are expressed as code.
%
CoCoEvo \cite{li2025cocoevo} simultaneously evolves programs and test cases from natural language descriptions. 

%
However, no prior work has explored the potential of evolutionary CAD code generation with LLMs.
In this work, we propose and investigate a method for generating CAD objects through CAD code using VLMs and evolutionary optimization.

\subsection{LLM-based CAD Code Generation}
The use of symbolic programs, like CAD code, for generating visual data is as old as computer graphics itself \cite{ritchie2023neurosymbolic}.
Historically designed as an efficient and accurate medium, CAD code allows LLMs to generate CAD objects, which unlocks new possibilities in product development and engineering. 

Popular CAD codes are PythonOCC \cite{paviot2022pythonocc}, which provides features for 3D CAD modeling with Python, and OpenSCAD \cite{openscad}, that a is a well-known CAD software that has been the leading scripting language for creating CAD objects.
However, many recent papers that work with LLMs and CAD code have chosen CADQuery \cite{cadquery2024}, as it follows a design-intended approach and is built in Python, making it more suitable for LLM applications \cite{alrashedy2025cadprompt}.

%
%
%
%
In the early work of \cite{picard2023concept}, the authors evaluate VLMs on multiple engineering tasks. Given a technical drawing, they prompted a VLM to generate OpenSCAD code, but their experiments were limited to a single object.
Similarly, \cite{yuan2024openecad} utilize a VLM to generate PythonOCC code from sketches. CAD-Coder \cite{doris2025cadcoder} reconstructs CAD objects from images by generating CADQuery code.
However, these methods rely on sketch or image inputs, which restricts their applicability and generalizability compared to using free text input.
\cite{ocker2025ideaCAD} presented an LLM-based multi-agent architecture for CAD, inspired by established development processes, that can work with both sketches and text prompts. Yet, similar to many previously mentioned methods, their approach still requires human feedback in a validation loop.


3D-Premise \cite{yuan2023-3dpremise} removes the need for user intervention by providing a VLM additional refinement prompts in the loop to correct any discrepancies.
CADCodeVerify \cite{alrashedy2025cadprompt} further improves the self-refinement by introducing an automatic feedback algorithm. Their method first generates a question catalog about the object using the prompt, in a second step the VLM utilizes these questions for refinement. 
They also introduce the first CAD code generation benchmark dataset CADPrompt, which we use in our experiments. 

While these prior works remove the need for human intervention, they are solely refinements of a single LLM answer. As a result, they do not utilize the full exploratory capabilities of LLMs, resulting in no observable improvement beyond the second refinement \cite{alrashedy2025cadprompt}.
In this work, we present a method utilizing the explorative power of evolutionary computation for generating CAD code from text prompts.

\label{chapter:related_work}

\section{Method}
In this section, we first formally define the problem and then introduce our method for generating CAD code using LLMs and evolutionary optimization.
Our method first generates an initial set $C$ of CAD code. In a second step, the initial set is further optimized. Fig.~\ref{fig:overview} showcases a high-level visual illustration of our method.

\subsection{Problem Definition}
%
%
We formally define a generation model $f(p)$, that takes a prompt $p$ and outputs CAD code $c$. The prompt describes an object $O$ via natural language and the CAD code can be converted into a 3D object $\hat{O}=\phi(c)$. The goal is to find a generation model $f$ such as the generated object $\hat{O}$ is geometrically and semantically close to its target $O$.
Note that the target object is not considered to be available during the procedure, and the generation is solely based on the text prompt.

\subsection{EvoCAD}
Our generation model follows an evolutionary strategy by first sampling an initial population, followed by an optimization, which can be classically divided into evaluation, crossover and mutation. The following outlines these steps in greater detail.

\textbf{Initialization.} During initialization, a language model $\mathrm{LM}$ takes the prompt $p$ and $k$ randomly selected few-shot samples from the CADQuery documentation. The $\mathrm{LM}$ is asked to generate CADQuery code for the object described in $p$. This process is repeated $M$ times, producing a diverse initialization set due to a non-zero temperature of the $\mathrm{LM}$ and varying few-shot samples. 
If a generated code is not compilable, we apply one attempt of self-debugging following the strategy outlined in \cite{chen2024codefix} by providing the output error message to the $\mathrm{LM}$.

\textbf{Evaluation.} In the evaluation phase, the fitness of the generated objects is determined. 
Usually a mathematical function evaluates the fitness of samples in the population. A common choice are distance functions that measure how much a candidate solution deviates from an optimum or ideal value.
However, during generation, we need to assume to not have access to a ground truth object $O$, making the calculation of distance metrics infeasible.
This reduces the problem down to a text-to-3D alignment, i.e., how well the object follows the description in the prompt. Recent work \cite{wu2024gptv-eval} has shown that VLMs, such as GPT-4V \cite{openai2023gpt4}, can evaluate 3D objects with nearly human level quality.

In our method, the fitness evaluation is a two-step process. First, a vision-language model $\mathrm{VLM}$ is prompted to generate a textual description of each object using chain-of-thought prompting \cite{wei2022chain}.
In the second step, a reasoning language model $\mathrm{RLM}$ takes these object descriptions, carefully compares them to the prompt and to each other, and finally ranks the objects from best alignment to the prompt to worst.
In each evaluation, the ranking procedure is performed three times, and the average object rankings are calculated.
We have intentionally divided the evaluation process into two distinct steps to fully leverage the strengths of each language model. VLMs are great at visual analysis, while RLMs excel at making well-considered decisions following complex reasoning.

\textbf{Crossover.} 
After transforming the average rankings into an exponential probability distribution, the resulting probabilities for each object are used to select mating pairs through a weighted random choice.
Equation \ref{eq:exp_distr} illustrates the distribution used, where $r_i$ represents the average rank of an object $O_i$, $R$ denotes the set of rankings, and $\lambda$ is the parameter of the exponential distribution.
\begin{equation} \label{eq:exp_distr}
\begin{split}
p(r_i,\:R,\:\lambda) = \frac{ e^{-\lambda\:r_i}}{\sum_{r_j \in R} e^{-\lambda\:r_j}}
\end{split}
\end{equation}
During the mating process, we give the $\mathrm{LM}$ the CAD codes of the two objects, their respective descriptions, and the prompt.
Provided with this information, we ask the model to analyze similarities, strengths and weaknesses of the respective CAD codes, and to combine them in a complementary way to generate an improved version. In this way, two objects are mated and an offspring object is generated.

\textbf{Mutation.} 
Additionally, with a probability of $p_m$, some offspring are selected for mutation. If selected, we provide the offspring's CAD code and prompt to the $LM$ and request that the model refine and improve the code.
In this way, new and potentially enhanced procedural representations can evolve.

Algorithm \ref{alg:method} outlines our method in detail.
During initialization a set $C$ containing $M$ CAD codes is generated (line 1-5). 
To obtain a diverse set $C$, $k$ few-shots samples are randomly selected (line 3) and in addition to the prompt $p$ provided to a $LM$, which generates a CAD object through code (line 4). 
After initialization, $N$ generations of evolutionary optimization are conducted (line 6-16). 
Each CAD code is rendered into a multiview image (line 8), which is subsequently analyzed by a $VLM$ (line 9). 
The generated textual descriptions, along with the prompt, are then provided to the $RLM$, which ranks the objects from best to worst alignment based on the information contained in the prompt (line 10).
The ranking process is independently repeated three times, and the resulting rankings are averaged (line 11).
After the selection of parent pairs (line 12), the previously described crossover is performed (line 13), followed by mutation (line 14). Finally, the population is updated (line 15).

\SetKwComment{Comment}{\#}{}
\SetKwInput{KwData}{Given}
\RestyleAlgo{ruled} 
\LinesNumbered

\begin{algorithm}
\caption{EvoCAD Algorithm}
\label{alg:method}
\SetAlgoLined
\KwData{prompt $p$, language model $\mathrm{LM}$, vision language model $\mathrm{VLM}$, reasoning language model $\mathrm{RLM}$, few-shot samples $S$, number of few-used shot samples $k$, population size $M$, number of generations $N$, mutation probability $p_m$, renderer $\phi$}
\BlankLine
\SetKwFunction{FFitness}{evaluate}
\SetKwProg{Fn}{Function}{:}{end}
\tcp{Initialize Population}
 $C \gets  [\:]$\;
\For{$i \gets 1$ \KwTo $M$}{
    $s \gets \mathrm{randomSelection}(S,\:k)$\;
    $C_i \gets \mathrm{LM}(p,\:s_i)$\;
} 
\BlankLine
\tcp{Evolutionary Optimization}
\For{$gen \gets 1$ \KwTo $N$}{

    \tcp{Generate description}
    $\mathrm{Images},\:\mathrm{Descriptions} \gets  [\:]$\;
    $\mathrm{Images} \gets \phi(C)$\;
    $\mathrm{Descriptions} \gets \mathrm{VLM}(\mathrm{Images})$\;

    \tcp{Object ranking}
    $R_1,\:R_2,\:R_3\gets \mathrm{RLM}(p,\:\mathrm{Descriptions})$\;
     $R_{avg} \gets \mathrm{avg}(R_1, R_2, R_3) $\;
    \tcp{Crossover}
    $P \gets \mathrm{selectParents}(R_{avg},\:C)$\;
    $C' \gets \mathrm{LM}(P,\:\mathrm{Descriptions},\:p)$\;
    \tcp{Mutation}
    $C'' \gets \mathrm{LM}(C',\:p_m,\:p)$\;
    $C \gets \mathrm{update}(C, C', C'')$\;
}

\end{algorithm}



\label{chapter:method}

\section{Experiments}
In this section, we evaluate our evolutionary approach for generating CAD objects with VLMs.

\subsection{Experimental Setup}
Experiments are conducted using GPT-4V \cite{openai2023gpt4} as generating $\mathrm{LM}$ and visually analyzing $\mathrm{VLM}$, allowing a fair comparison to the prior methods 3D-Premise \cite{yuan2023-3dpremise} and CADCodeVerify \cite{alrashedy2025cadprompt}, which are based on the same model. 
Performance data for these methods is obtained from \cite{alrashedy2025cadprompt}.
Additionally, we evaluate our method using GPT-4o \cite{openai2024gpt4ocard} to assess the applicability of our approach to the ongoing trend of enhancing VLMs.

In all our experiments, we chose the recent reasoning model o3-mini \cite{openai2025gpto3-mini} as the evaluating $\mathrm{RLM}$, balancing complex reasoning and efficiency. 
For each model, GPT-4V and GPT-4o, we conduct three independent runs. Unless otherwise stated, the depicted results show the average values calculated across these three runs.
Our method is applied on the full CADPrompt dataset \cite{alrashedy2025cadprompt}, which consists of 200 samples in total. 

Similar to \cite{alrashedy2025cadprompt}, we set the number of few-shot samples $k$ to 5 for initialization by leveraging a set $S$ of 40 samples from the CADQuery documentation.
We chose the initial population size $M$ to be 6, and the number of optimization generations $N$ to be 4. 
A mutation rate $p_m$ of 50\% encourages offspring to regularly explore alternative novel solution strategies. The exponential probability distribution parameter from Eq. \ref{eq:exp_distr} is set to $\lambda=0.5$.
Additionally, we chose the number of elites to be one, i.e., we move the object with the highest fitness unmodified to the next generation, ensuring that the current best object does not deteriorate.
The LLMs temperature is set to 0.2 for evaluation tasks, such as visual object description and ranking, and to 0.5 for all other generative operations, including initialization, crossover, and mutation.
This configuration enables a more objective evaluation while enhancing creativity in the generation process.

\subsection{Evaluation Metrics}\label{subsec:eval}
Point cloud distance (PCD) and Hausdorff distance (HDD) compare the geometric properties between the generated object $\hat{O}$ and the ground truth $O$, and are used in previous works \cite{yuan2023-3dpremise, alrashedy2025cadprompt}. In the following $d(o, \hat{o})$, denotes the Euclidean distance between the points $o$ and $\hat{o}$, which lie on the surface of objects $O$ and $\hat{O}$ respectively.
\begin{equation} \label{eq:pcd}
\begin{split}
PCD(O,\:\hat{O}) = \frac{1}{2|O|}\sum_{o \in O}\min_{\hat{o} \in \hat{O}} d(o,\hat{o})   +  \frac{1}{2|\hat{O}|} \sum_{{\hat{o}}\in {\hat{O}}}\min_{{o} \in O} d(o, \hat{o})
\end{split}
\end{equation}
\begin{equation} \label{eq:hdd}
\begin{split}
HDD(O,\:\hat{O}) = \max \{ \sup_{o\in O}\inf_{\hat{o} \in \hat{O}} d(o,\hat{o}),\: \sup_{\hat{o} \in \hat{O}}\inf_{o \in O} d(o, \hat{o}) \}
\end{split}
\end{equation}
While PCD only utilizes the object surfaces, intersection-based metrics consider the volume overlap. 
Therefore, we utilize two popular intersection metrics: Intersection over Union (IoU) and the Dice coefficient (DSC).
\begin{equation} \label{eq:iou}
\begin{split}
IoU(O,\:\hat{O}) = \frac{|O\cap\hat{O}|}{|O\cup\hat{O}|}
\end{split}
\end{equation}
\begin{equation} \label{eq:dsc}
\begin{split}
DSC(O,\:\hat{O}) = \frac{2|O\cap\hat{O}|}{|O|+|\hat{O}|}
\end{split}
\end{equation}


Despite providing a broad indication of spatial alignment, these metrics do not capture semantic similarity.
For this reason, developing a more granular metric capable of capturing structural differences between objects remains an open research question \cite{alrashedy2025cadprompt}.
%
However, unlike text or images, the semantics of a texture-less CAD object must be characterized solely by its geometry.

A simple but elegant characterization choice is the topological property of an object defined by the Euler characteristic $\chi$, which is a topological invariant for 2-manifold polyhedra, e.g., watertight triangle meshes. 
It was originally stated by Euler in 1758 \cite{euler1758} and can be efficiently computed with $\chi=V-E+F$, where $V$, $E$ and $F$ are the numbers of vertices, edges, and faces respectively. Intuitively, $\chi$ informs about the number of ``holes'' in an object. Given an object $O$ with $g$ holes, the Euler characteristic is $\chi(O)=2-2g$. 

We argue that this characterization is particularly well-suited for CAD objects, which typically constitute mechanical components that have complex structures with multiple holes and cavities. 
In this manner, we are introducing two novel metrics: the topology error ($T_{err}$), which indicates the Euler characteristic error between $O$ and $\hat{O}$, and the topology correctness ($T_{corr}$), which informs whether the Euler characteristic of the generated object matches that of the ground truth. 
The latter is specifically designed to indicate the percentage of samples that are topologically correct when evaluating performance within a larger dataset.

\begin{equation} \label{eq:terr}
\begin{split}
T_{err} = |\chi(O)-\chi(\hat{O})|
\end{split}
\end{equation}
\begin{equation} \label{eq:tcorr}
\begin{split}
T_{corr} = \mathds{1}_{\chi(O)}(\chi(\hat{O}))
\end{split}
\end{equation}

To enable a fair quantitative comparison, all metrics depict the average values calculated over the same watertight joint subset, which approximately represents 80\% of the full dataset. 
Before calculating the spatial metrics PCD, HDD, IoU, and DSC, normalization is performed, followed by the iterative closest point (ICP) method to optimally align the generated object with the ground truth.
%



\subsection{Results}
\def\tcorrpremise{0.7987421383647799}
\def\tcorrcadprompt{0.8050314465408805}

\def\terrpremise{0.5786163522012578}
\def\terrcadprompt{0.6289308176100629}

\begin{figure*}
\begin{center}
    \begin{tikzpicture}
        \begin{groupplot}[
        group style={
            group name=my plots,
            group size=2 by 1,
            xlabels at=edge bottom,
            ylabels at=edge left,
            horizontal sep=1.5 cm,
            vertical sep=1.5cm,
            },
        xlabel={Number of Generation},
        label style={font=\footnotesize},
        tick label style={font=\footnotesize},
        legend style={font=\footnotesize},
        legend cell align=left,
        legend style={
            legend columns=4
        },
        xticklabel style={
            /pgf/number format/fixed,
            /pgf/number format/precision=4
        },
        scaled x ticks=false,
        width=\textwidth / 2,
        height=\textheight / 3.8,]

    \nextgroupplot[title=Topology Correctness (\%) $\uparrow$, legend to name=plot_legend, ymin=75, ymax=88.5, ylabel=Average $T_{corr}$]
    \addplot[name path=oursgptv, color=colorgpto, mark=x, mark size=1.5pt, thick]     coordinates {
        (0, 82.97654281838518)
        (1, 84.34722510194203)
        (2, 86.26315210665929)
        (3, 87.12792633436365)
        (4, 87.1938732856454)
    };
    \addlegendentry{EvoCAD-4o}

    \addplot[name path=oursgptv, color=colorgptv, mark=x, mark size=1.5pt, thick] 
    coordinates {
        (0, 78.30198611968862)
        (1, 79.3935335125679)
        (2, 81.20561843508568)
        (3, 82.10099982436275)
        (4, 82.44059980517535)
    };
    \addlegendentry{EvoCAD-4v}

    \addplot[domain=0:4, color=colorcadprompt, thick, dashed] {\tcorrcadprompt*100};
    \addlegendentry{CADCodeVerify}

    \addplot[domain=0:4, color=colorpremise, thick, dashed] {\tcorrpremise*100}; 
    \addlegendentry{3D-Premise}

    \addplot [name path=gpto_corr_up, draw=none]   
    coordinates {
        (0, 82.97654281838518+1.289)
        (1, 84.34722510194203+0.676)
        (2, 86.26315210665929+0.065)
        (3, 87.12792633436365+0.695)
        (4, 87.1938732856454+0.372)
    };
    \addplot [name path=gpto_corr_down, draw=none]   
    coordinates {
        (0, 82.97654281838518-1.289)
        (1, 84.34722510194203-0.676)
        (2, 86.26315210665929-0.065)
        (3, 87.12792633436365-0.695)
        (4, 87.1938732856454-0.372)
    };
    \addplot [colorgpto, fill opacity=0.2] fill between[of=gpto_corr_up and gpto_corr_down];
    
    \addplot [name path=gptv_corr_up, draw=none]   
    coordinates {
        (0, 78.30198611968862+2.578)
        (1, 79.3935335125679+2.009)
        (2, 81.20561843508568+2.757)
        (3, 82.100999824362751+2.238)
        (4, 82.44059980517535+2.312)
    };
    \addplot [name path=gptv_corr_down, draw=none]   
    coordinates {
        (0, 78.30198611968862-2.578)
        (1, 79.3935335125679-2.009)
        (2, 81.20561843508568-2.757)
        (3, 82.100999824362751-2.238)
        (4, 82.44059980517535-2.312)
    };
    \addplot [colorgptv, fill opacity=0.2] fill between[of=gptv_corr_up and gptv_corr_down];

    \nextgroupplot[title=Topology Error $\downarrow$, ymin=0.36, ymax=0.71, ylabel=Average $T_{err}$]
    \addplot[name path=oursgptv, color=colorgpto, mark=x, mark size=1.5pt, thick]     coordinates {
        (0, 0.5005314494770654)
        (1, 0.47202524938374)
        (2, 0.4151823184009643)
        (3, 0.39611268412600265)
        (4, 0.4052891821506862)
    };
    
    \addplot[name path=oursgptv, color=colorgptv, mark=x, mark size=1.5pt, thick]     coordinates {
        (0, 0.6379037783033344)
        (1, 0.5740015282329377)
        (2, 0.49536895309647805)
        (3, 0.4753938091340977)
        (4, 0.45420227007624625)
    };

    \addplot[domain=0:4, color=colorcadprompt, thick, dashed] {\terrcadprompt};

    \addplot[domain=0:4, color=colorpremise, thick, dashed] {\terrpremise};

    \addplot [name path=gpto_err_up, draw=none]   
    coordinates {
        (0, 0.5005314494770654+0.048)
        (1, 0.47202524938374+0.02)
        (2, 0.4151823184009643+0.003)
        (3, 0.39611268412600265+0.019)
        (4, 0.4052891821506862+0.013)
    };
    \addplot [name path=gpto_err_down, draw=none]   
    coordinates {
        (0, 0.5005314494770654-0.048)
        (1, 0.47202524938374-0.02)
        (2, 0.4151823184009643-0.003)
        (3, 0.39611268412600265-0.019)
        (4, 0.4052891821506862-0.013)
    };
    \addplot [colorgpto, fill opacity=0.2] fill between[of=gpto_err_up and gpto_err_down];
    
    \addplot [name path=gptv_err_up, draw=none]   
    coordinates {
        (0, 0.6379037783033344+0.061)
        (1, 0.5740015282329377+0.049)
        (2, 0.49536895309647805+0.1)
        (3, 0.4753938091340977+0.087)
        (4, 0.45420227007624625+0.069)
    };
    \addplot [name path=gptv_err_down, draw=none]   
    coordinates {
        (0, 0.6379037783033344-0.061)
        (1, 0.5740015282329377-0.049)
        (2, 0.49536895309647805-0.1)
        (3, 0.4753938091340977-0.087)
        (4, 0.45420227007624625-0.069)
    };
    \addplot [colorgptv, fill opacity=0.2] fill between[of=gptv_err_up and gptv_err_down];

    \end{groupplot}
    \end{tikzpicture}
    \ref{plot_legend}
\caption[Plots]{Left: $T_{corr}$ defined as the ratio of samples where the Euler characteristic of the generated object matches that of the ground truth. Right: $T_{err}$ as average difference between the Euler characteristics of the generated object and the ground truth.\label{fig:plot}}
\end{center}
\end{figure*}
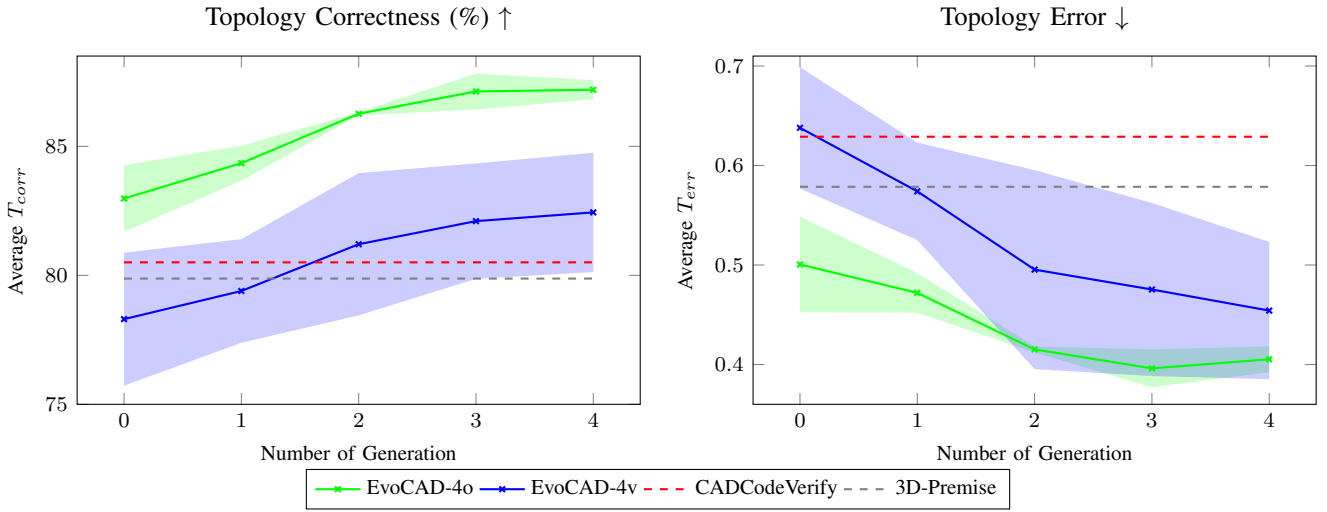



\begin{table*}[hbt]
\centering
\caption{Quantitative comparison of methods on the CADPrompt benchmark \cite{alrashedy2025cadprompt}. \textbf{Bold} entries indicate the best results, with standard deviations from three repeated runs shown in parentheses. See section \ref{subsec:eval} for metric details.}
\label{table:quantitative}
    \begin{tabular}{lllllll}
    \toprule
     Method  &   $T_{corr}\:(\%)\:\uparrow$ &  $T_{err}\:\downarrow$  & $PCD\:\downarrow$ & $HDD\:\downarrow$ & $IoU\:(\%)\:\uparrow$ & $DSC\:(\%)\:\uparrow$ \\
    \midrule
     3D-Premise~\cite{yuan2023-3dpremise} & 79.9              &    0.579        &    0.0660        & 0.189                 & 68.2              & 77.6  \\
     CADCodeVerify~\cite{alrashedy2025cadprompt}      & 80.5             &     0.629         &  0.0628         & 0.182                & 69.8     &  79.0  \\
    EvoCAD-4v     & 82.4 (2.3)    & 0.446 (0.069)    & 0.0626 (0.0007)    & 0.180 (0.002)      &  69.7 (0.4)                 &  79.1 (0.2)   \\

    EvoCAD-4o      &   \textbf{87.2} (0.4)  &  \textbf{0.410} (0.013)   &  \textbf{0.0617} (0.0020)         & \textbf{0.177} (0.004)            & \textbf{69.9} (1.0) &  \textbf{79.4} (0.8) \\
    \bottomrule
    \end{tabular}
\end{table*}

\textbf{Optimization.} 
Fig.~\ref{fig:plot} shows the optimization behavior per generation of our method using the two topology metrics $T_{corr}$ and $T_{err}$, which depict the semantic similarity of the generated object compared to the ground truth.
The shaded area around the line illustrates the standard deviation derived from the three independently conducted experiments.
Additionally, the performance of the two baseline methods, 3D-Premise \cite{yuan2023-3dpremise} and CADCodeVerify \cite{alrashedy2025cadprompt}, is included for reference.

Both plots demonstrate that our method effectively increases the topology correctness $T_{corr}$ and decreases the topological error $T_{err}$. 
While both metrics are inherently coupled, excelling in one does not necessarily lead to an improvement in the other. This is evident from the performance of the two baseline methods, where CADCodeVerify is superior to 3D-Premise in $T_{corr}$ but lags behind in $T_{err}$.

Initially, our method performs worse than the two baseline methods, which is an expected behavior. The baseline numbers are calculated on their final refined objects, whereas the objects in our initial generation are purely generated from the prompt without any feedback. However, our method is able to outperform the baselines after the second optimization generation in both metrics.
%
Despite the modern GPT-4o already surpassing the baselines in the initial generation, our method still enables further improvements.
%

\textbf{Quantitative evaluation.} 
Table~\ref{table:quantitative} demonstrates that our proposed approach significantly outperforms prior methods in the two topology metrics $T_{corr}$ and $T_{err}$.
%
Moreover, our method achieves the lowest PCD and HDD as well as the best DSC.
Only the IoU value of EvoCAD-4v is slightly lower than that of CADCodeVerify.
%
Specifically, EvoCAD-4o significantly outperforms both prior methods across all evaluation metrics.
One reason for this could be the advanced generation capabilities of GPT-4o. However, it also shows the effectiveness of our method when applied to modern VLMs.

Moreover, GPT-4o achieves more consistent results in topology metrics ($T_{corr}$, $T_{err}$), while GPT-4V shows greater consistency in the spatial metrics (PCD, HDD, IoU, DSC).
The reason for this may be a small negative correlation between the topology metrics and the spatial metrics.
We have frequently observed that objects with incorrect topologies can have better spatial metrics than those with correct topologies.
For example, missing a required hole is preferred to having one that deviates from the exact ground truth position.
While this introduces additional randomness into the generation process, it also highlights the value of our topology metrics, which capture quality properties not represented by other metrics.

\def\promptwidth{0.22}
\def\objectwidth{0.125}
\def\prompttextsize{\tiny}
\def\metrictextsize{\scriptsize}

\def\trima{0.06}
\def\trimb{0.03}
\def\trimc{0.08}
\def\trimd{0.04}
\def\trime{0.04}
\def\trimf{0.03}
\def\trimg{0.075}

\begin{figure*}
\centering
{
\begin{tabular}{cccccc} \\
Prompt & Ground Truth & EvoCAD-4o  & EvoCAD-4v & CADCodeVerify & 3D-Premise \\
\hline

\begin{minipage}{\promptwidth\textwidth}
\prompttextsize
\vspace*{-2.8 cm}
\input{02-images/00670454/00670454_prompt}
\vspace*{0.0 cm}
\end{minipage}
    &\adjincludegraphics[width=\objectwidth\textwidth, trim={{\trima\width} {\trima\width} {\trima\width} {\trima\width}},clip]{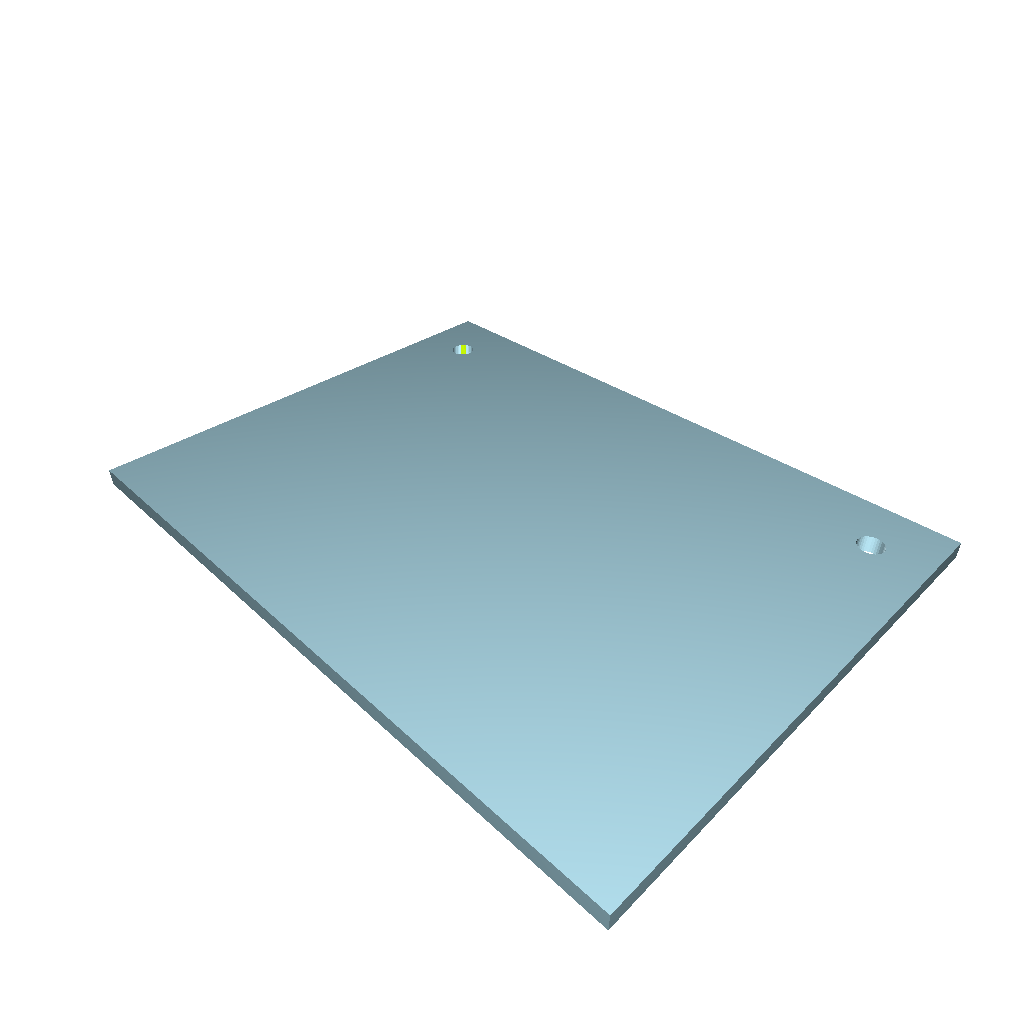}
   & \adjincludegraphics[width=\objectwidth\textwidth, trim={{\trima\width} {\trima\width} {\trima\width} {\trima\width}},clip]{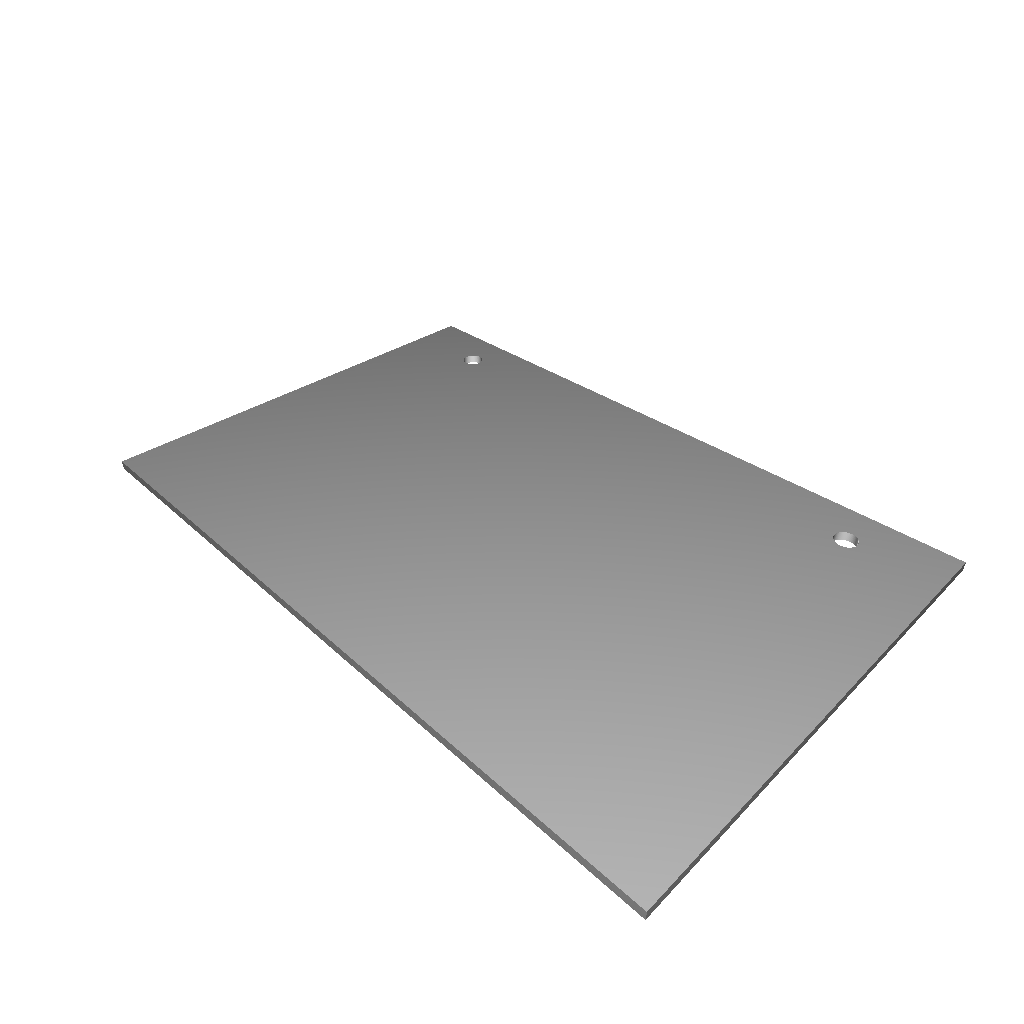} 
 & \adjincludegraphics[width=\objectwidth\textwidth, trim={{\trima\width} {\trima\width} {\trima\width} {\trima\width}},clip]{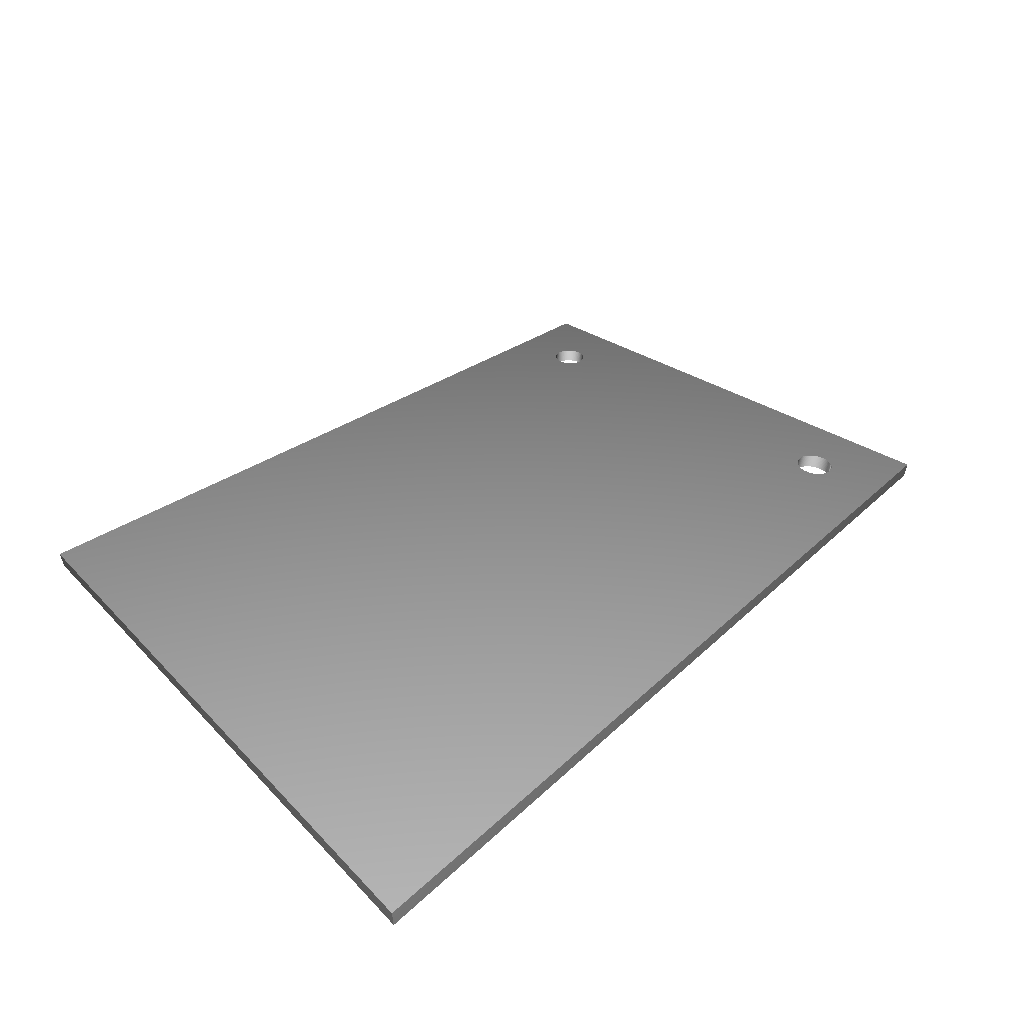} 
 & \adjincludegraphics[width=\objectwidth\textwidth, trim={{\trima\width} {\trima\width} {\trima\width} {\trima\width}},clip]{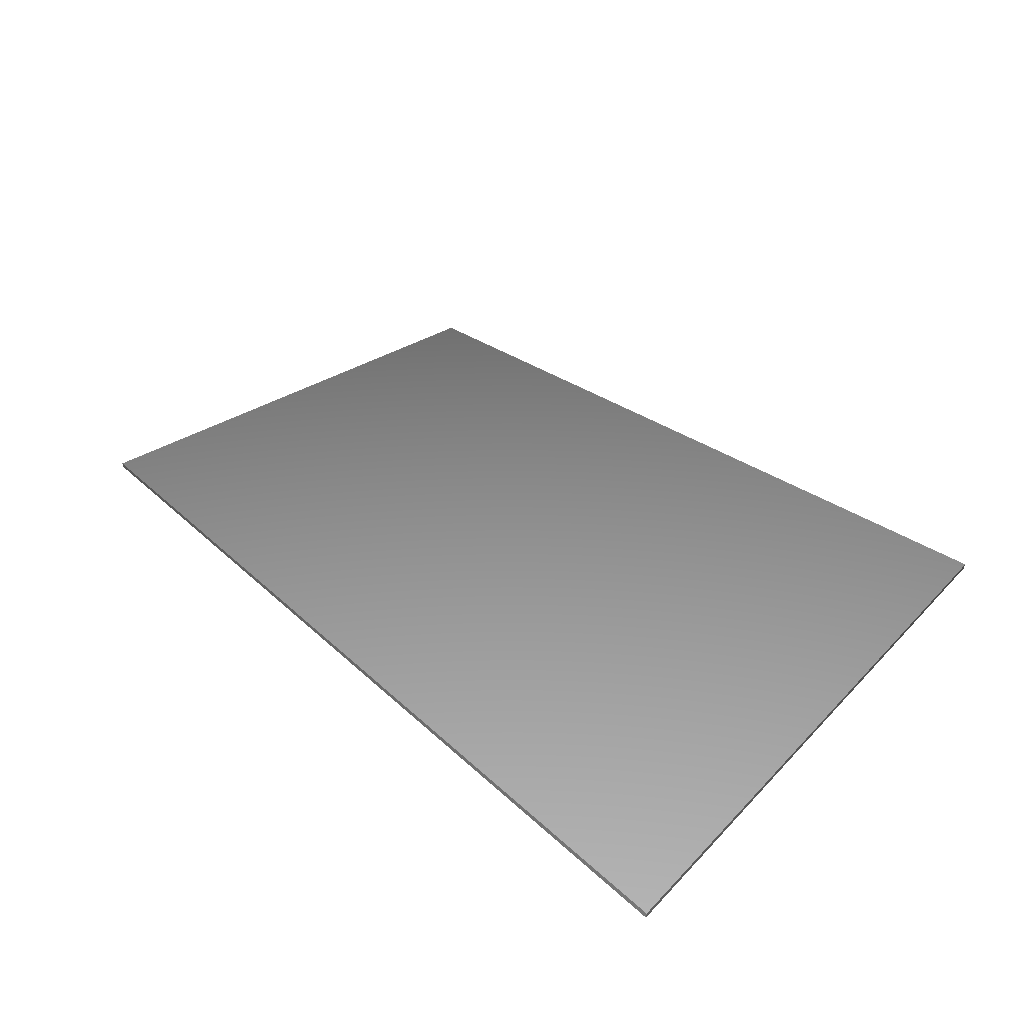} 
 & \adjincludegraphics[width=\objectwidth\textwidth, trim={{\trima\width} {\trima\width} {\trima\width} {\trima\width}},clip]{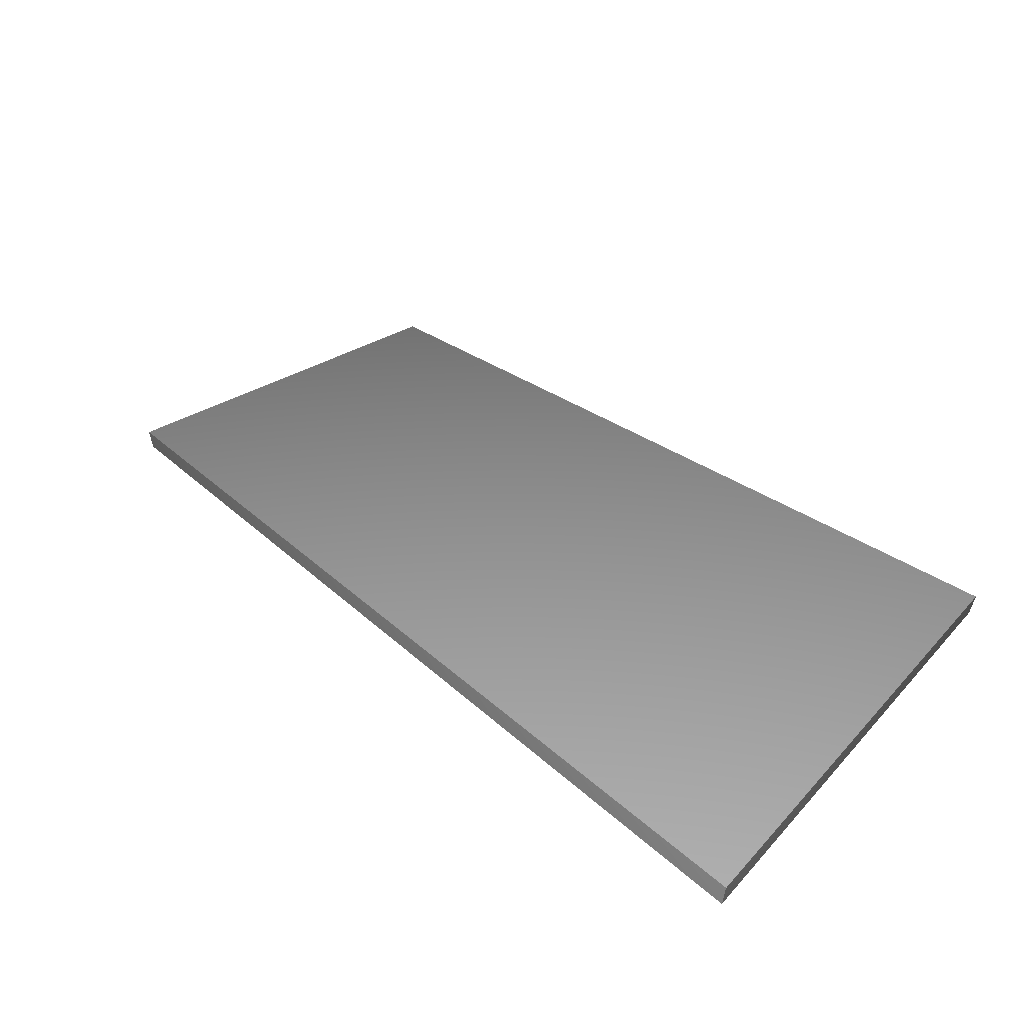}  \\
  & \metrictextsize $\chi=-2$   & \metrictextsize \textcolor{correct}{$\chi=-2$} & \metrictextsize \textcolor{correct}{$\chi=-2$} &  \metrictextsize \textcolor{false}{$\chi=2$} & \metrictextsize \textcolor{false}{$\chi=2$}  \\
\hline

\begin{minipage}{\promptwidth\textwidth}
\prompttextsize
\vspace*{-3.2 cm}
\input{02-images/00034239/00034239_prompt}
\vspace*{0.0 cm}
\end{minipage}
    &\adjincludegraphics[width=\objectwidth\textwidth, trim={{\trimb\width} {\trimb\width} {\trimb\width} {\trimb\width}},clip]{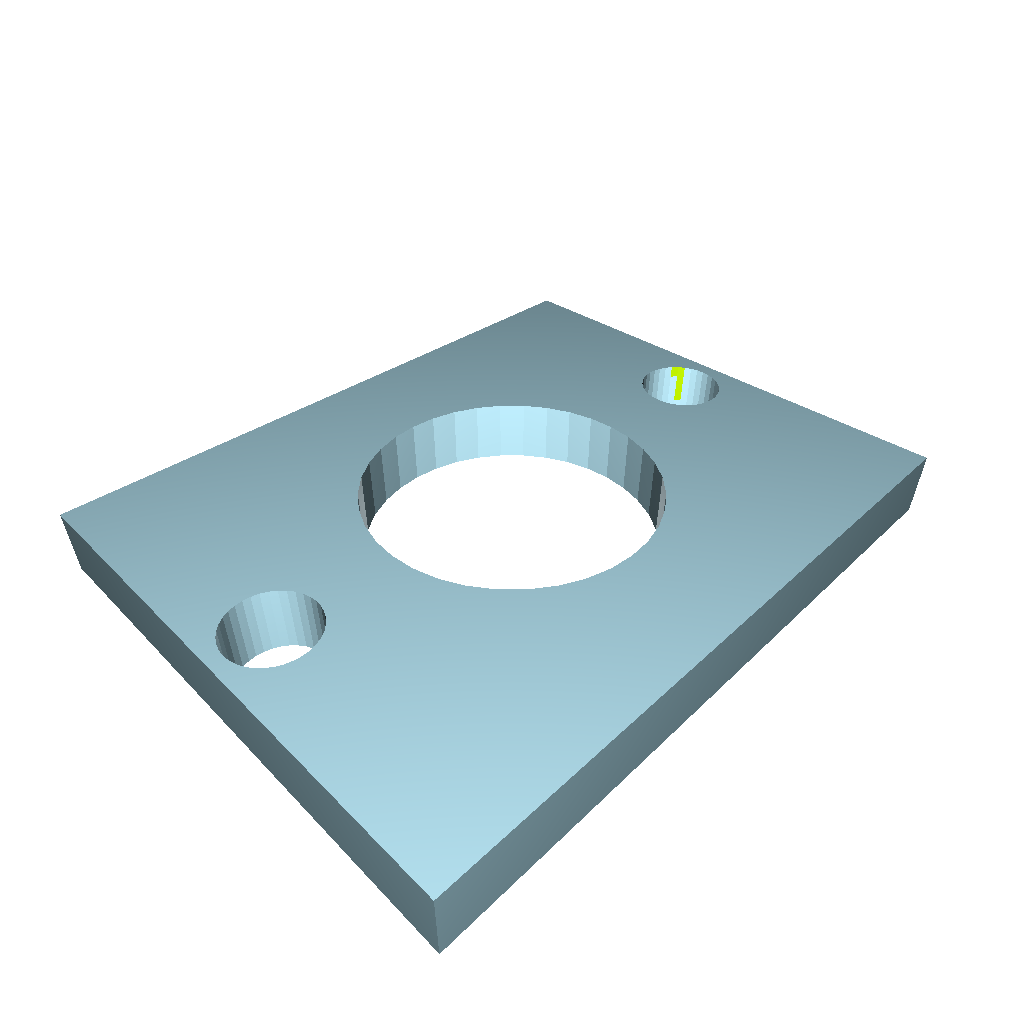}
   & \adjincludegraphics[width=\objectwidth\textwidth, trim={{\trimb\width} {\trimb\width} {\trimb\width} {\trimb\width}},clip]{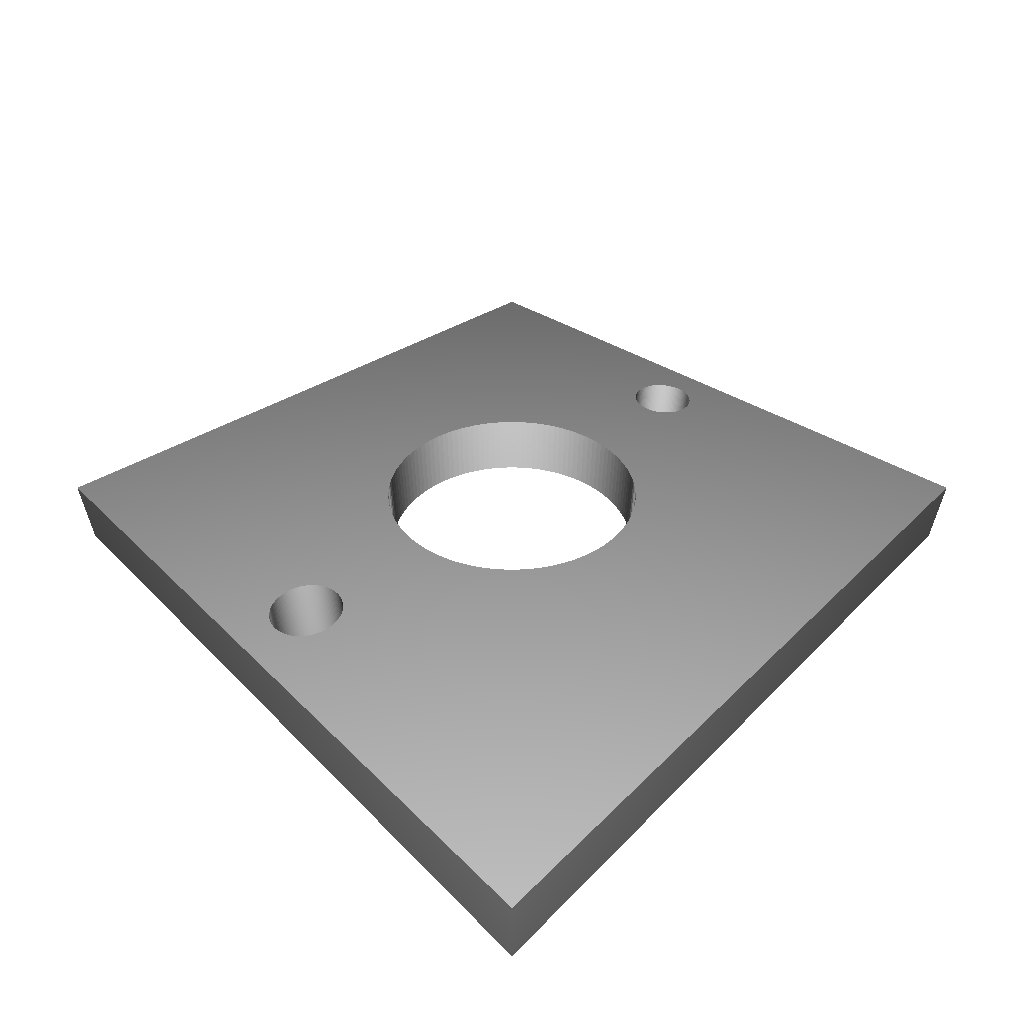} 
 & \adjincludegraphics[width=\objectwidth\textwidth, trim={{\trimb\width} {\trimb\width} {\trimb\width} {\trimb\width}},clip]{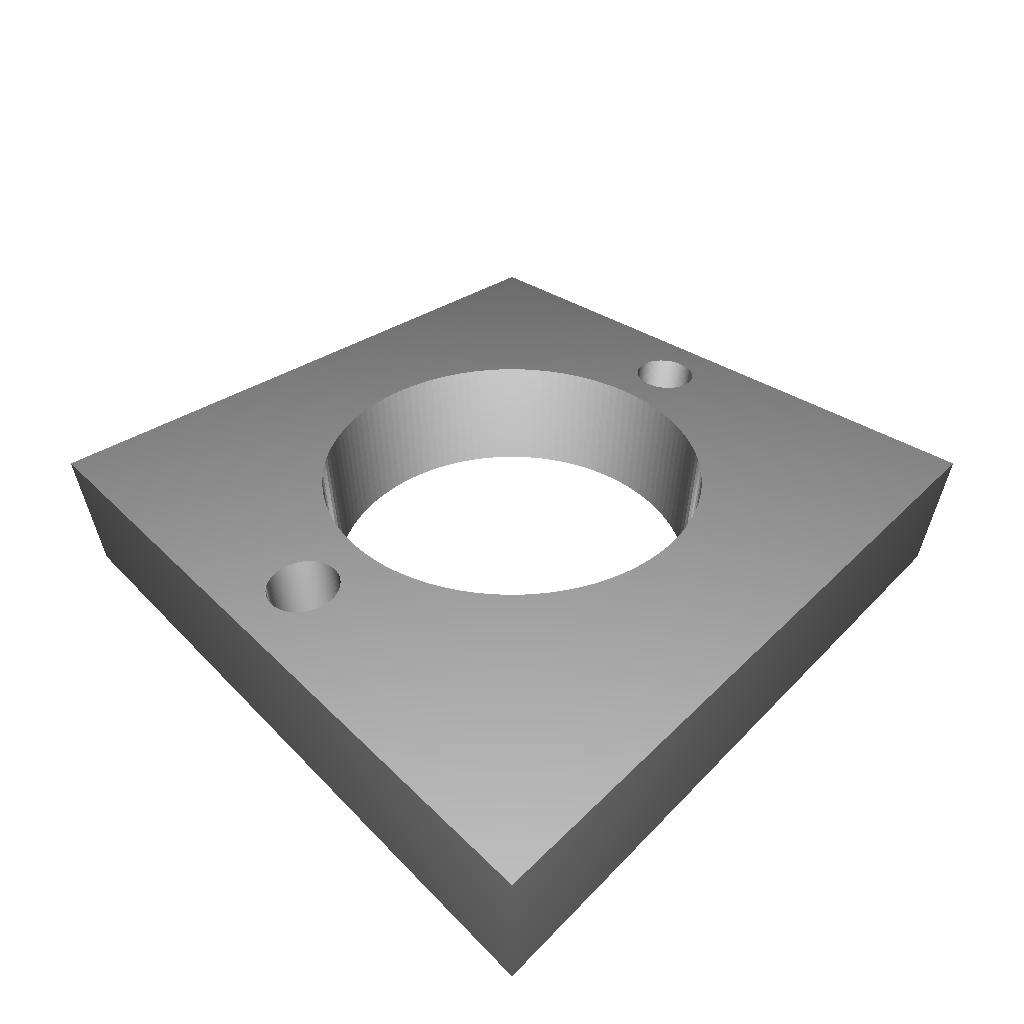} 
 & \adjincludegraphics[width=\objectwidth\textwidth, trim={{\trimb\width} {\trimb\width} {\trimb\width} {\trimb\width}},clip]{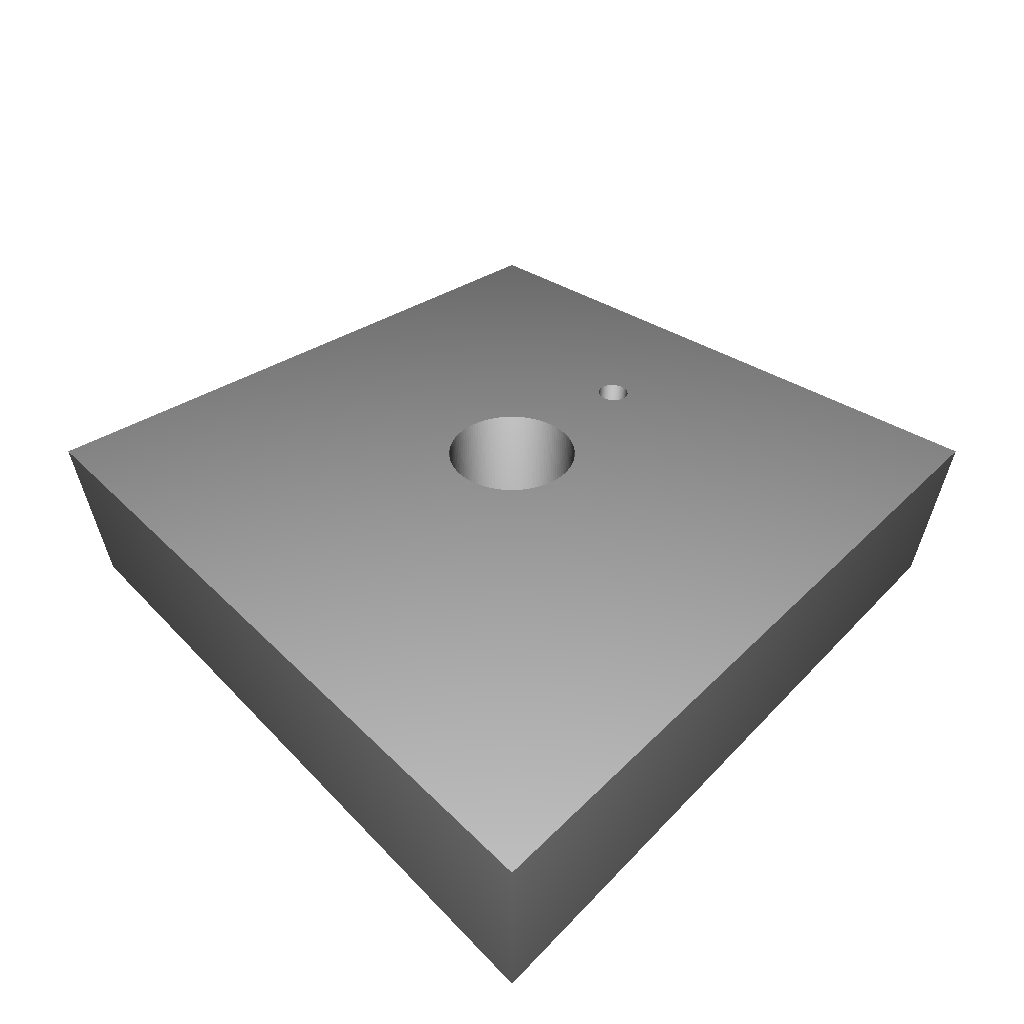} 
 & \adjincludegraphics[width=\objectwidth\textwidth, trim={{\trimb\width} {\trimb\width} {\trimb\width} {\trimb\width}},clip]{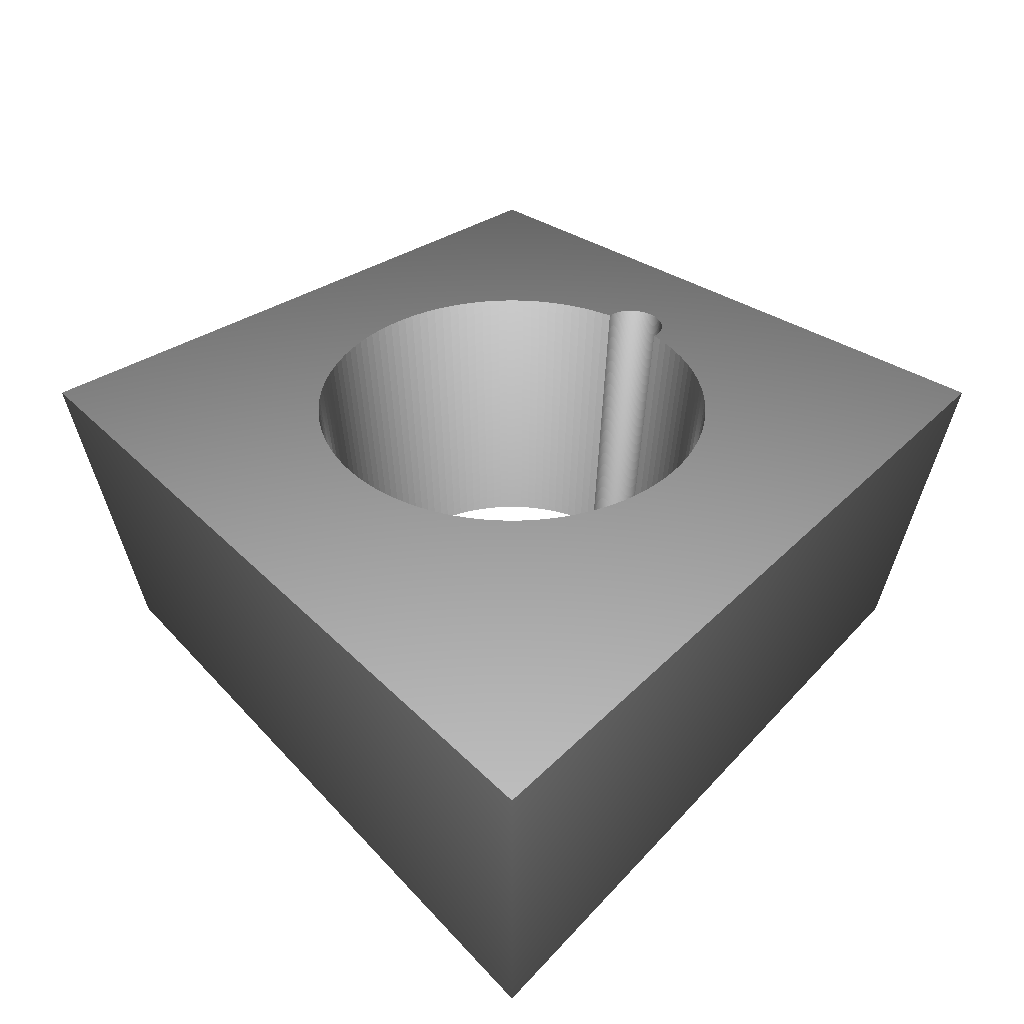} \\
  & \metrictextsize $\chi=-4$  & \metrictextsize \textcolor{correct}{$\chi=-4$} & \metrictextsize \textcolor{correct}{$\chi=-4$} &  \metrictextsize \textcolor{false}{$\chi=-2$} & \metrictextsize \textcolor{false}{$\chi=0$}  \\
\hline


\begin{minipage}{\promptwidth\textwidth}
\prompttextsize
\vspace*{-2.7 cm}
\input{02-images/00039365/00039365_prompt}
\vspace*{0.0 cm}
\end{minipage}
    &\adjincludegraphics[width=\objectwidth\textwidth, trim={{\trimd\width} {\trimd\width} {\trimd\width} {\trimd\width}},clip]{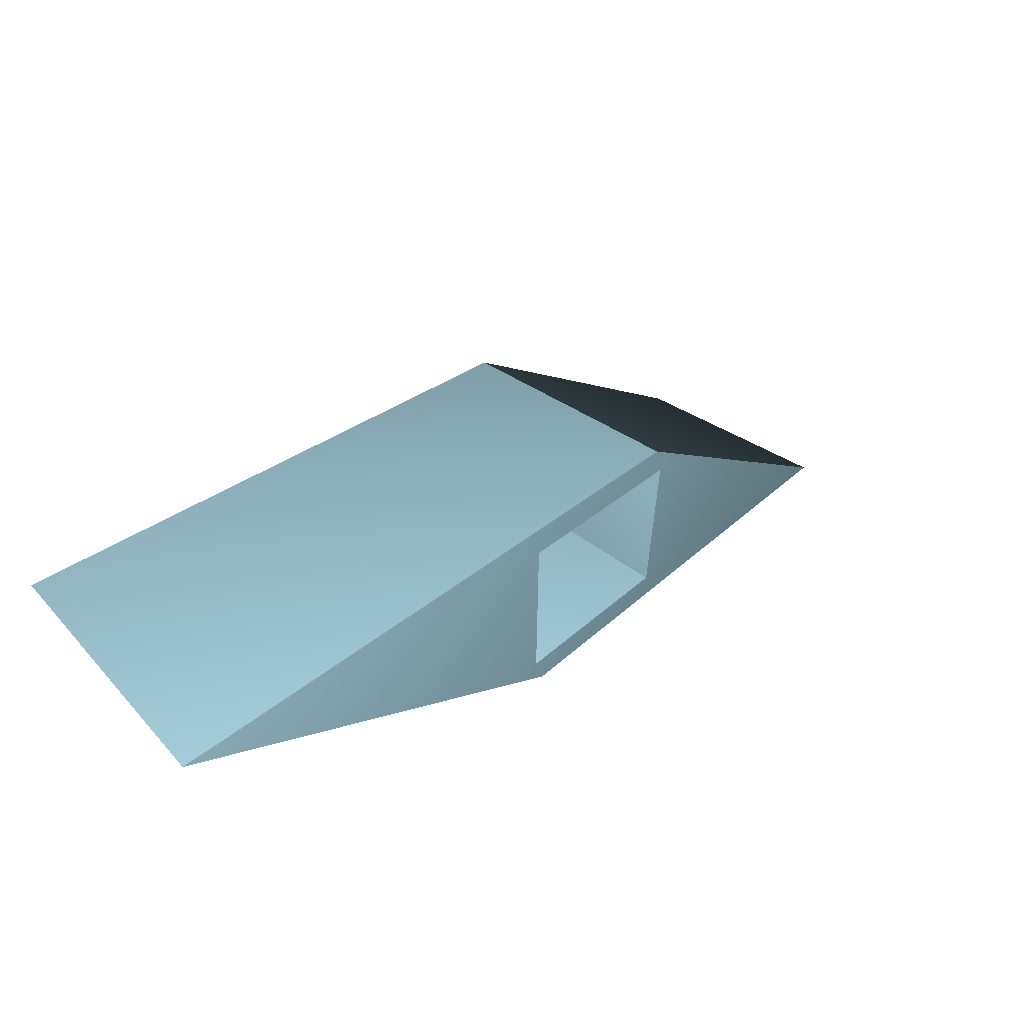}
   & \adjincludegraphics[width=\objectwidth\textwidth, trim={{\trimd\width} {\trimd\width} {\trimd\width} {\trimd\width}},clip]{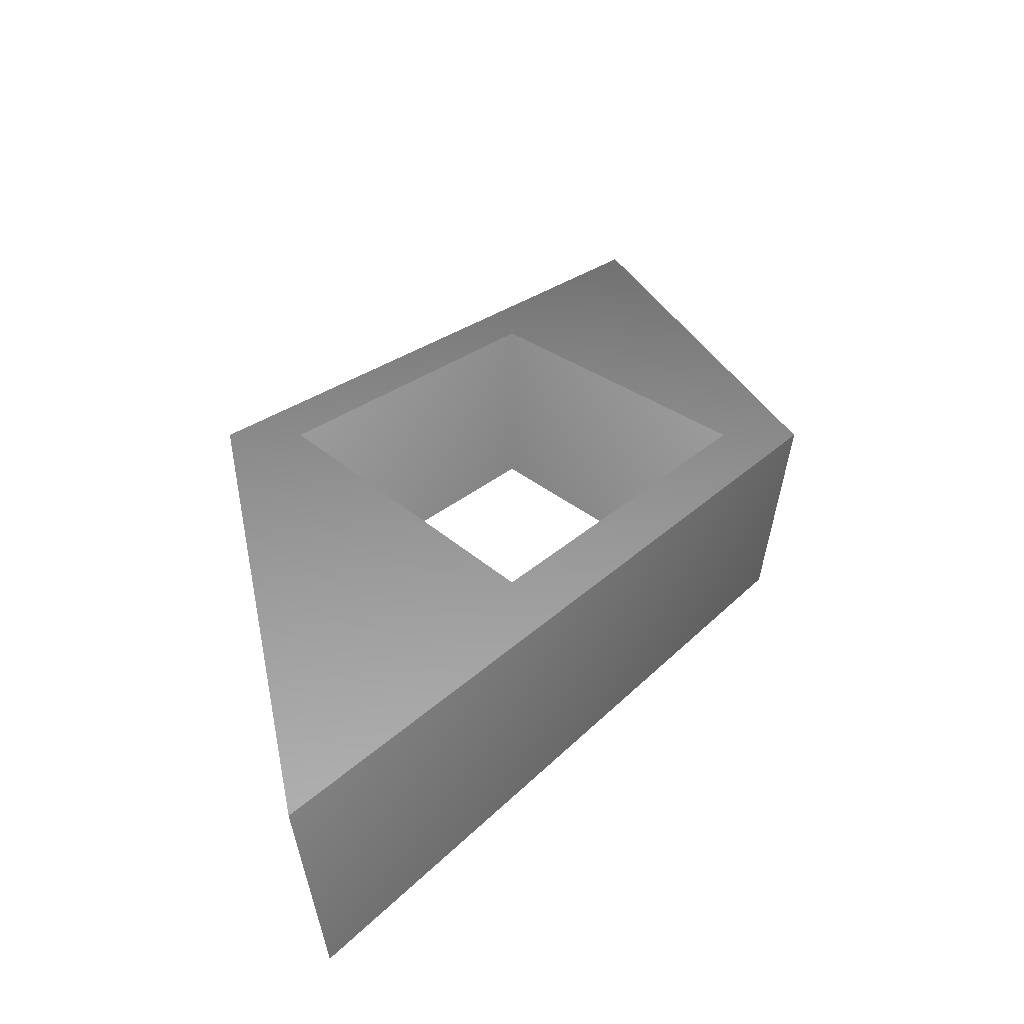}
 & \adjincludegraphics[width=\objectwidth\textwidth, trim={{\trimd\width} {\trimd\width} {\trimd\width} {\trimd\width}},clip]{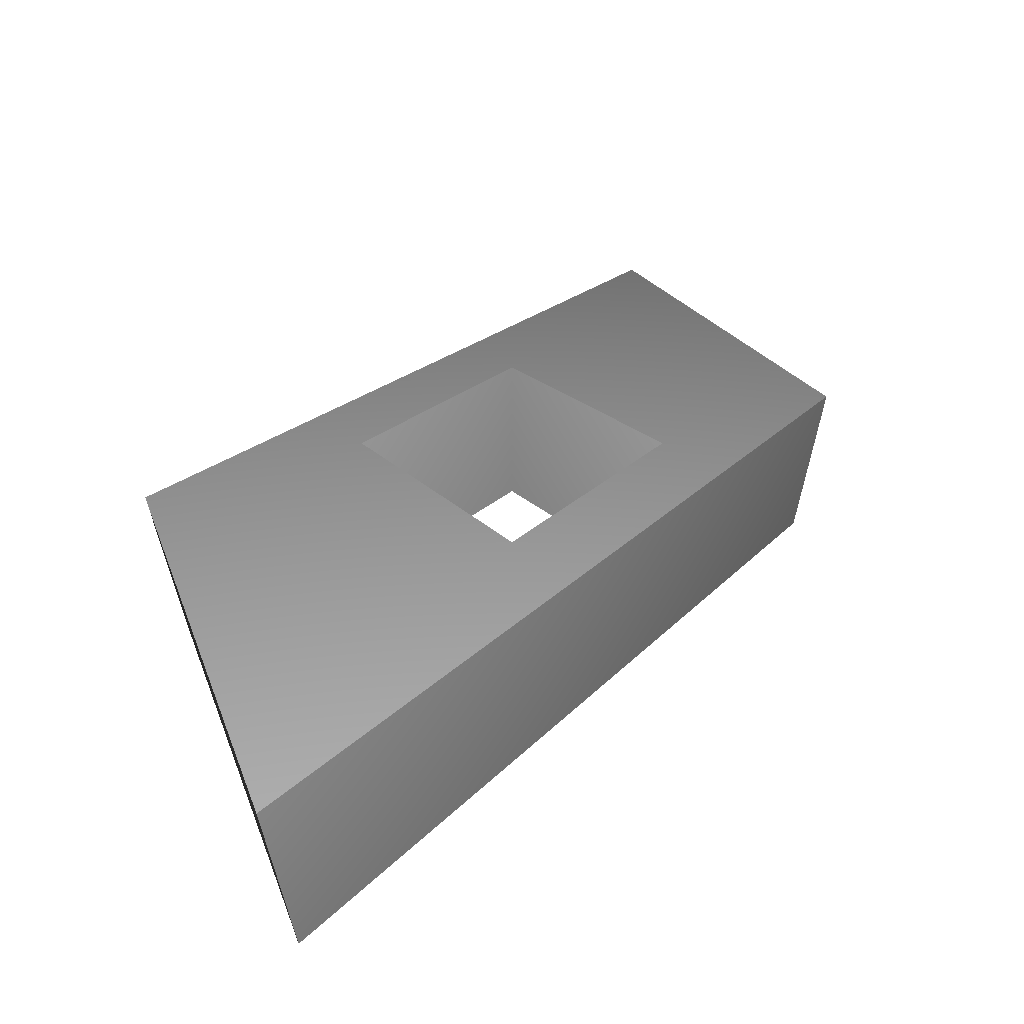} 
 & \adjincludegraphics[width=\objectwidth\textwidth, trim={{\trimd\width} {0.2\width} {\trimd\width} {\trimd\width}},clip]{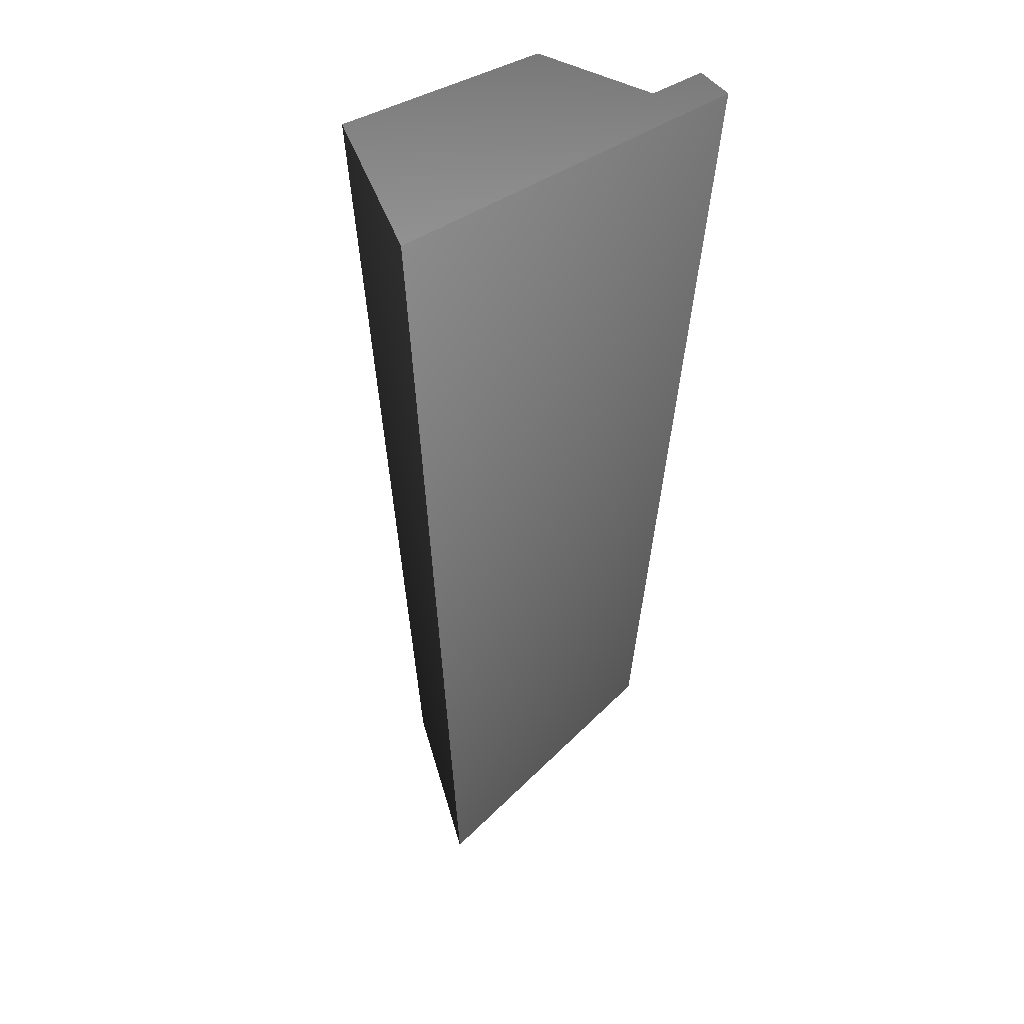} 
 & \adjincludegraphics[width=\objectwidth\textwidth, trim={{\trimd\width} {0.2\width} {\trimd\width} {\trimd\width}},clip]{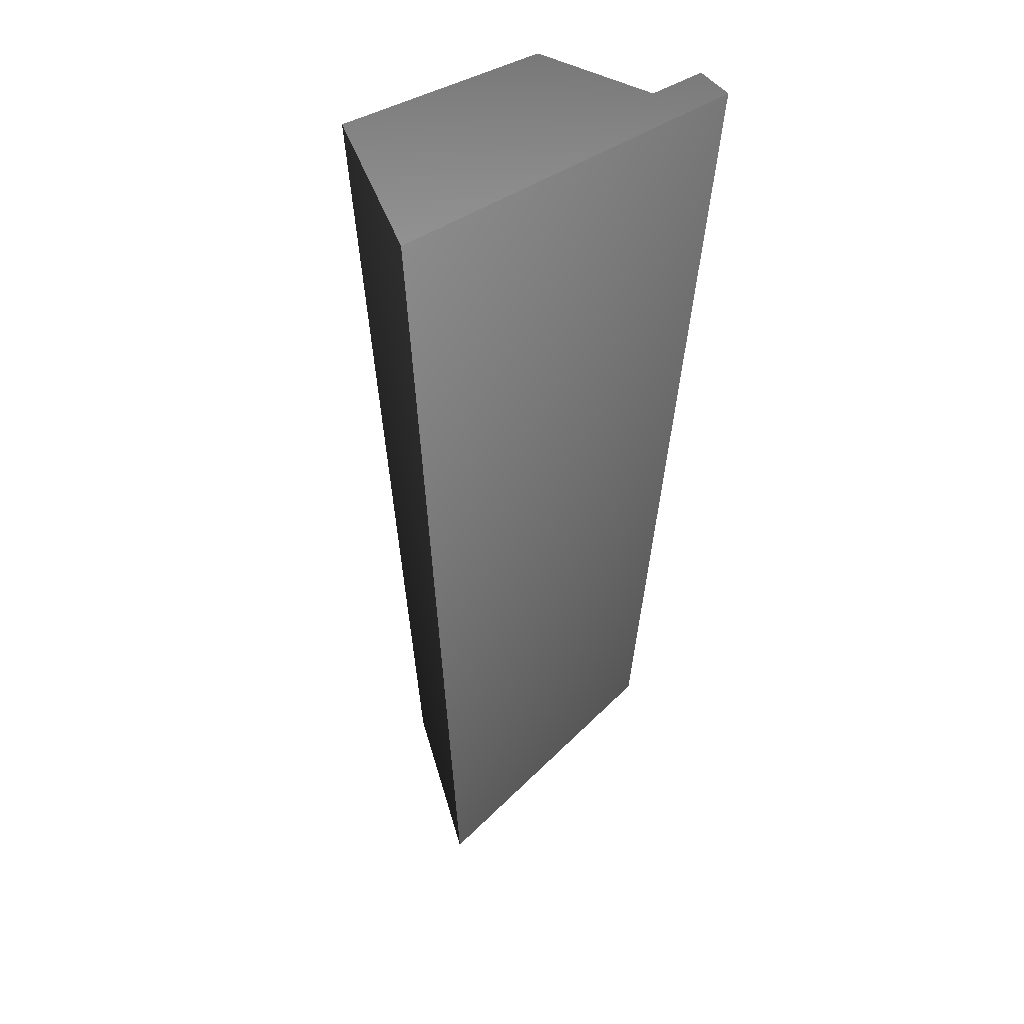}  \\
& \metrictextsize $\chi=0$  & \metrictextsize \textcolor{correct}{$\chi=0$} & \metrictextsize \textcolor{correct}{$\chi=0$} &  \metrictextsize \textcolor{false}{$\chi=2$} & \metrictextsize \textcolor{false}{$\chi=2$}  \\
\hline


\begin{minipage}{\promptwidth\textwidth}
\prompttextsize
\vspace*{-1.9 cm}
\input{02-images/00997065/00997065_prompt}
\vspace*{0.0 cm}
\end{minipage}
    &\adjincludegraphics[width=\objectwidth\textwidth, trim={{\trimc\width} {\trimc\width} {\trimc\width} {\trimc\width}},clip]{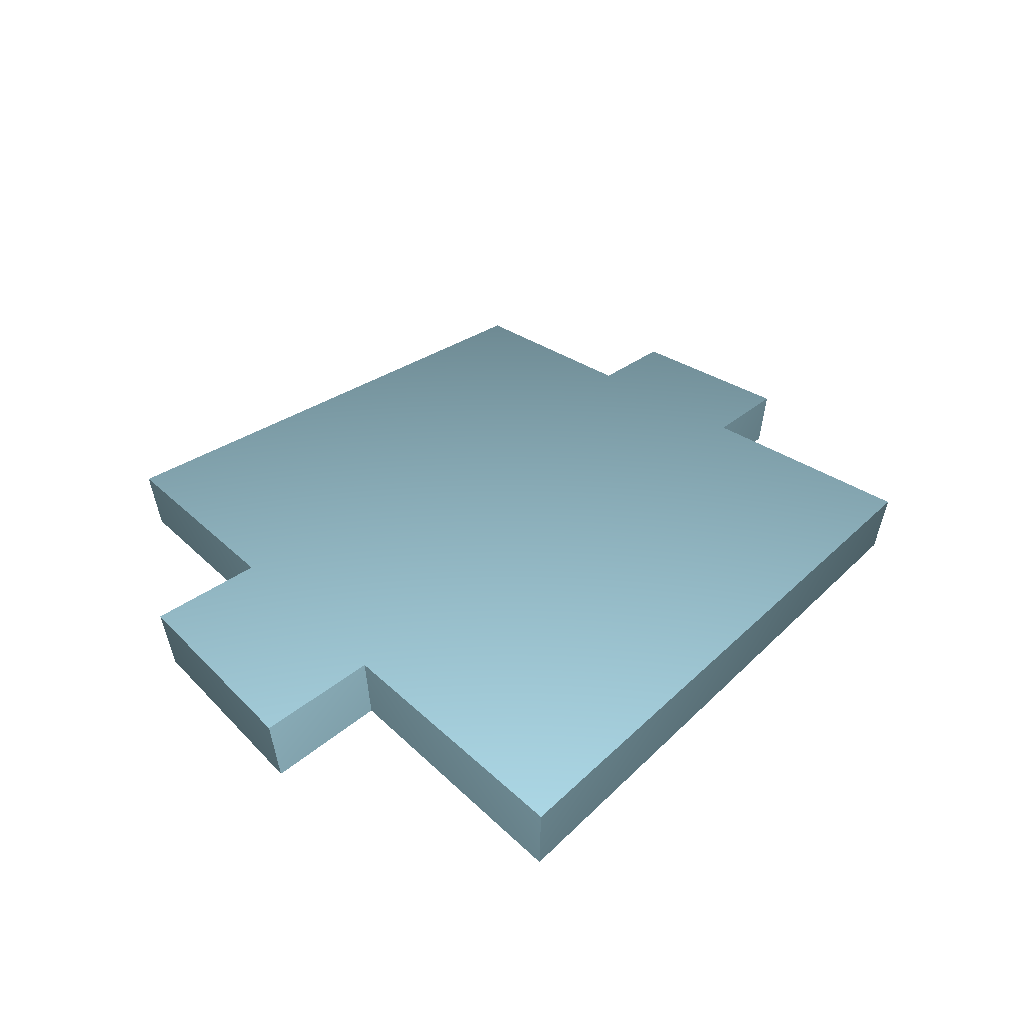}
     & \adjincludegraphics[width=\objectwidth\textwidth, trim={{\trimc\width} {\trimc\width} {\trimc\width} {\trimc\width}},clip]{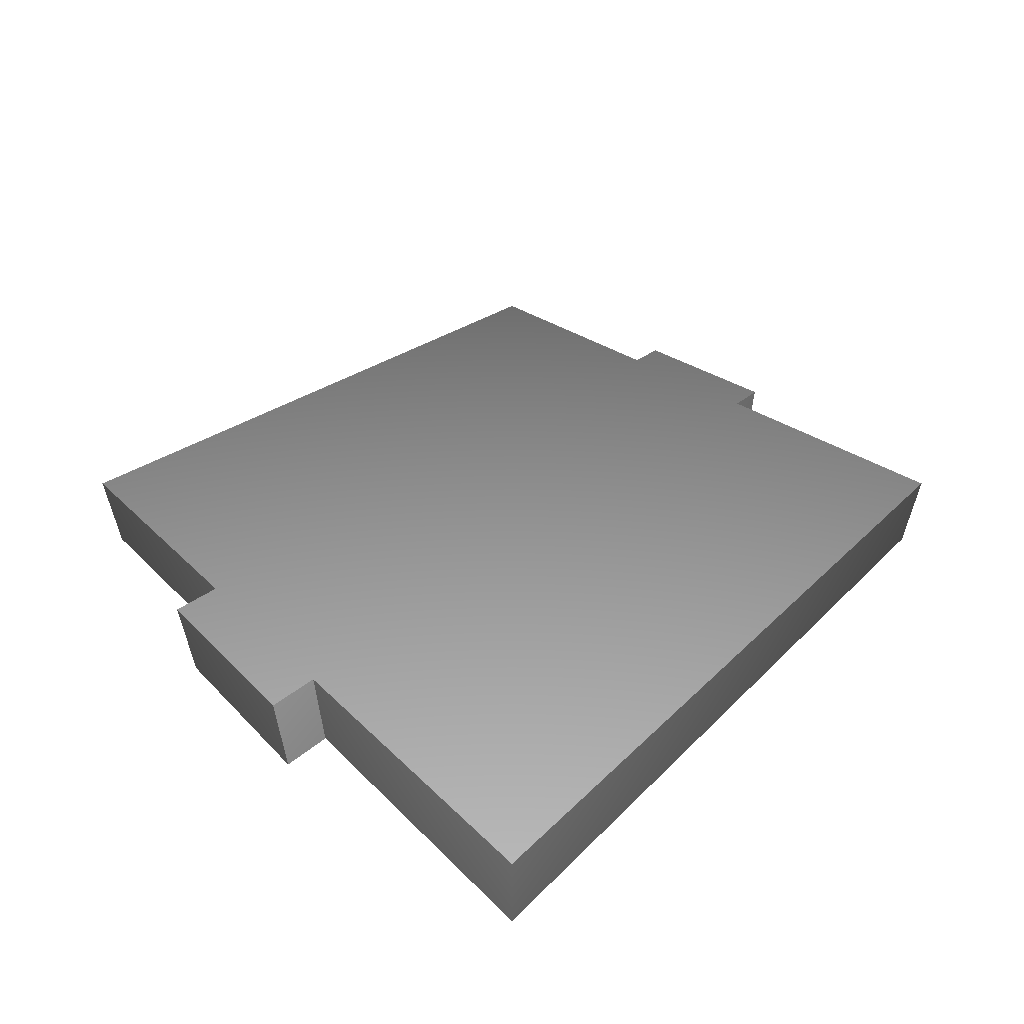} 
 & \adjincludegraphics[width=\objectwidth\textwidth, trim={{\trimc\width} {\trimc\width} {\trimc\width} {\trimc\width}},clip]{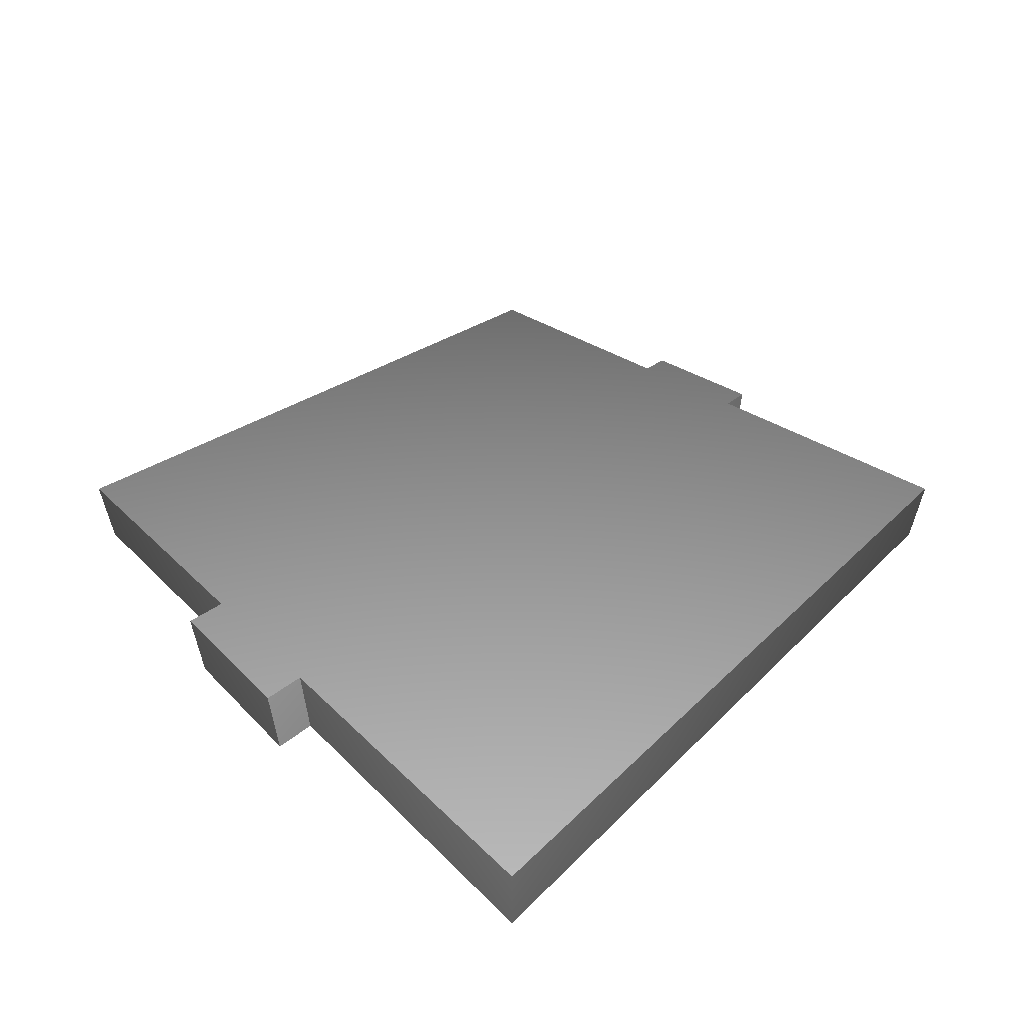} 
 & \adjincludegraphics[width=\objectwidth\textwidth, trim={{\trimc\width} {\trimc\width} {\trimc\width} {\trimc\width}},clip]{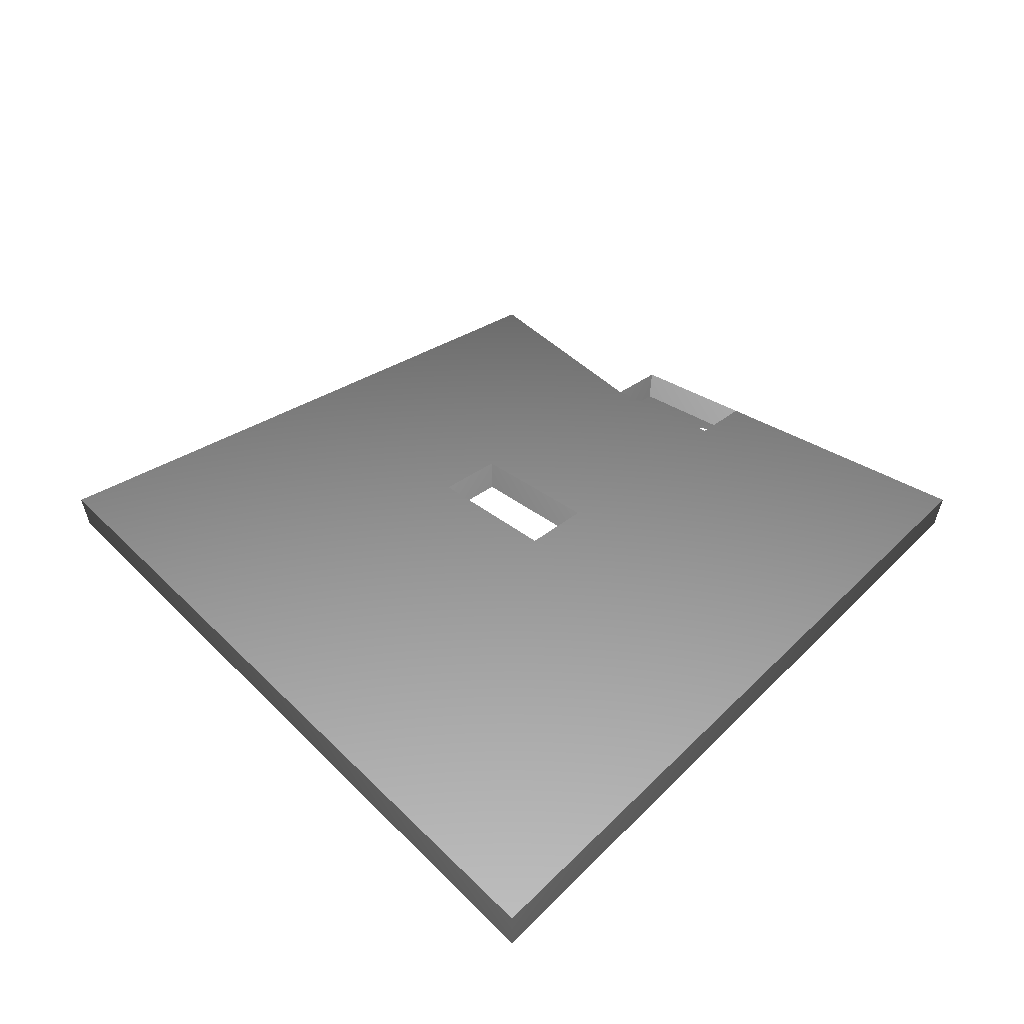} 
 & \adjincludegraphics[width=\objectwidth\textwidth, trim={{\trimc\width} {\trimc\width} {\trimc\width} {\trimc\width}},clip]{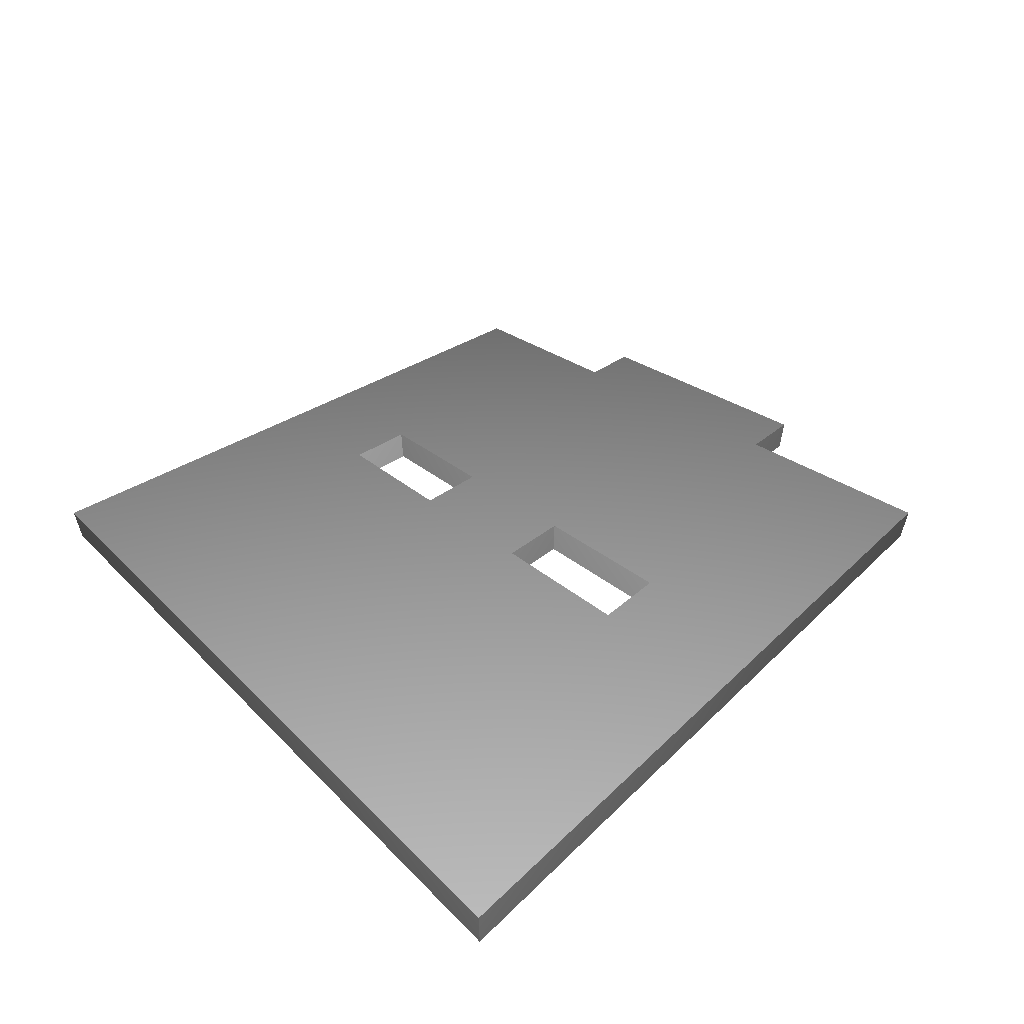}  \\
    & \metrictextsize $\chi=2$    & \metrictextsize \textcolor{correct}{$\chi=2$} & \metrictextsize \textcolor{correct}{$\chi=2$} &  \metrictextsize \textcolor{false}{$\chi=\mathrm{NaN}$} & \metrictextsize \textcolor{false}{$\chi=-2$}  \\
\hline

\begin{minipage}{\promptwidth\textwidth}
\prompttextsize
\vspace*{-2.0 cm}
\input{02-images/00682073/00682073_prompt}
\vspace*{0.0 cm}
\end{minipage}
    &\adjincludegraphics[width=\objectwidth\textwidth, trim={{\trime\width} {\trime\width} {\trime\width} {\trime\width}},clip]{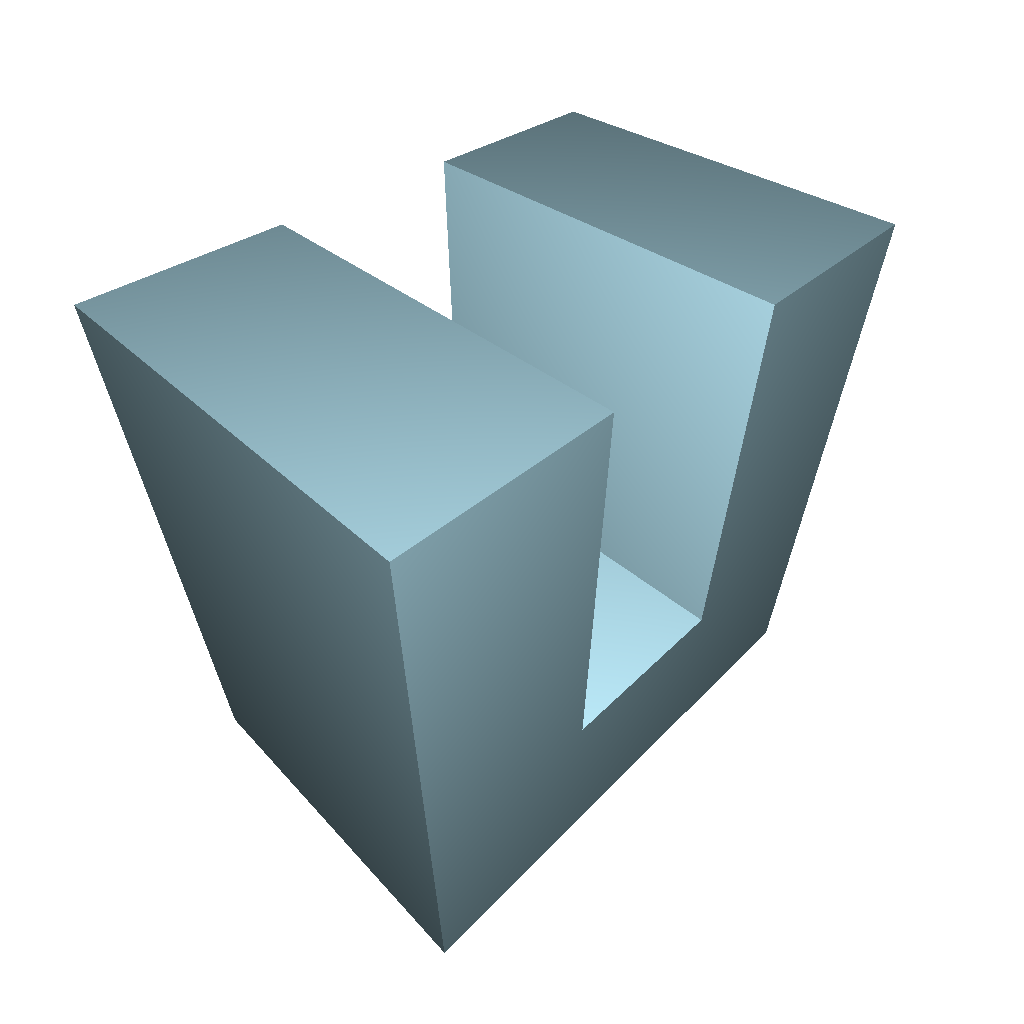}
  & \adjincludegraphics[width=\objectwidth\textwidth, trim={{\trime\width} {\trime\width} {\trime\width} {\trime\width}},clip]{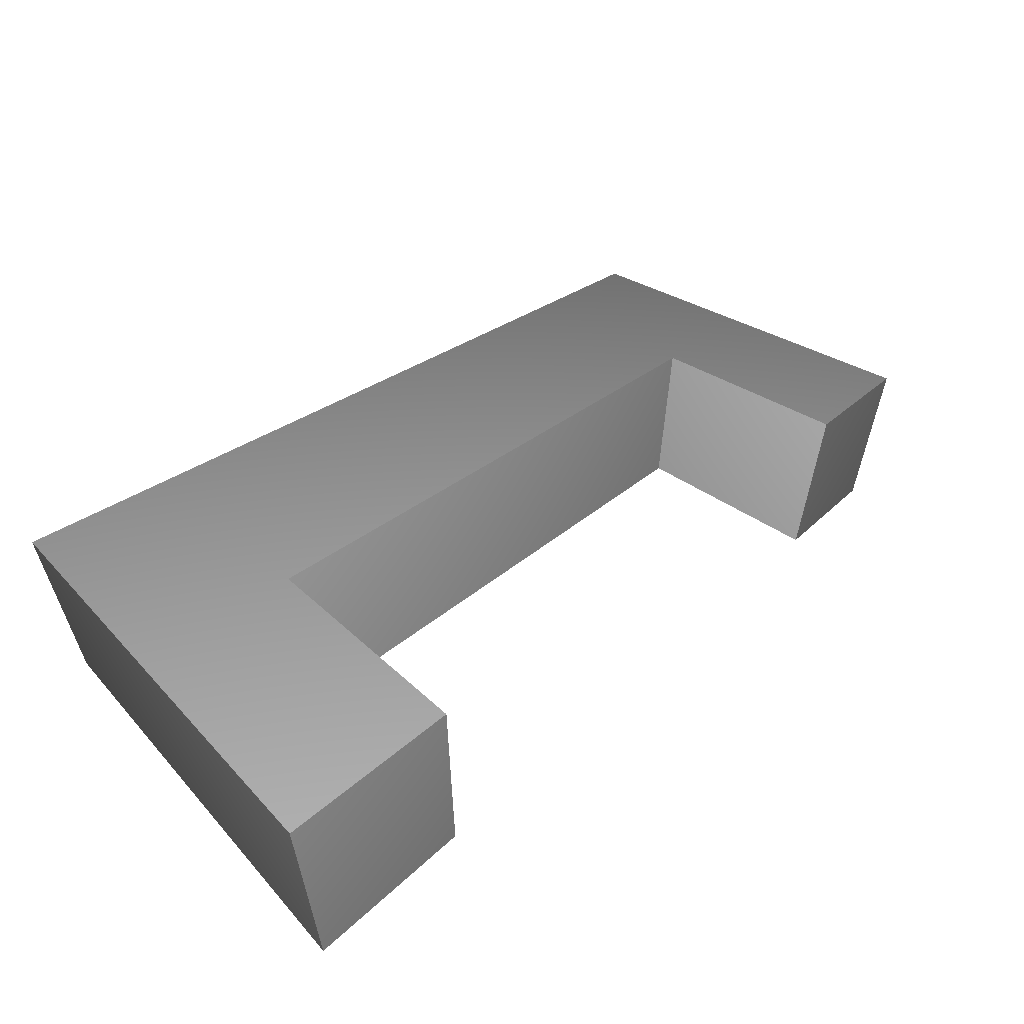} 
 & \adjincludegraphics[width=\objectwidth\textwidth, trim={{\trime\width} {\trime\width} {\trime\width} {\trime\width}},clip]{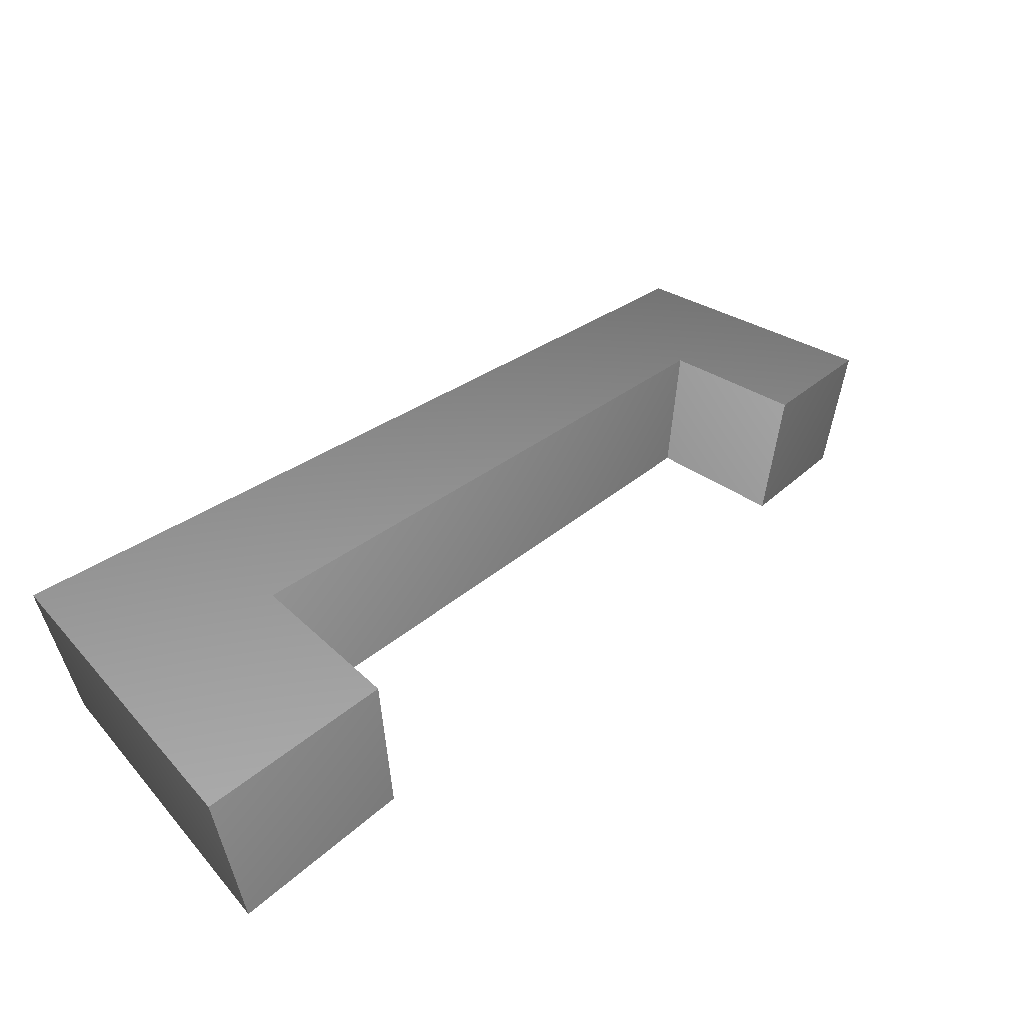} 
 & \adjincludegraphics[width=\objectwidth\textwidth, trim={{\trime\width} {\trime\width} {\trime\width} {\trime\width}},clip]{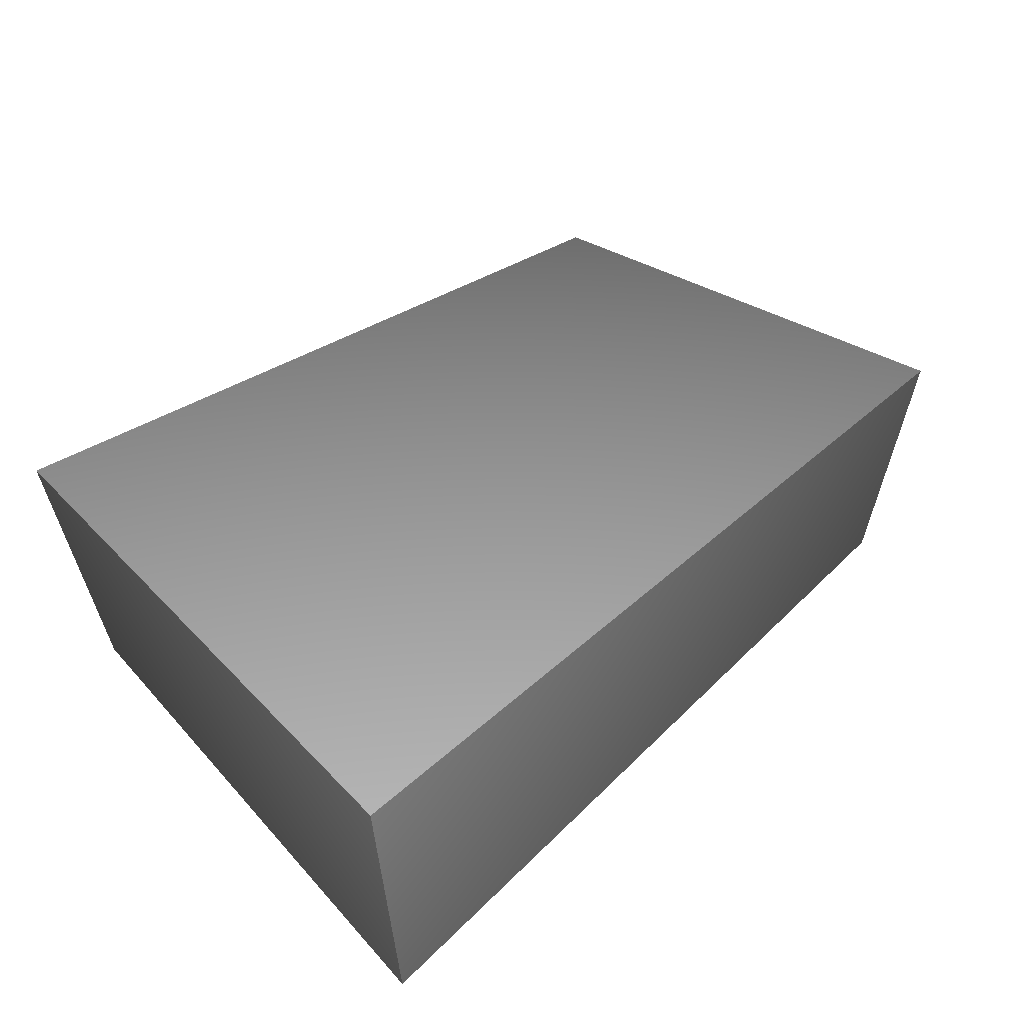} 
 & \adjincludegraphics[width=\objectwidth\textwidth, trim={{\trime\width} {\trime\width} {\trime\width} {\trime\width}},clip]{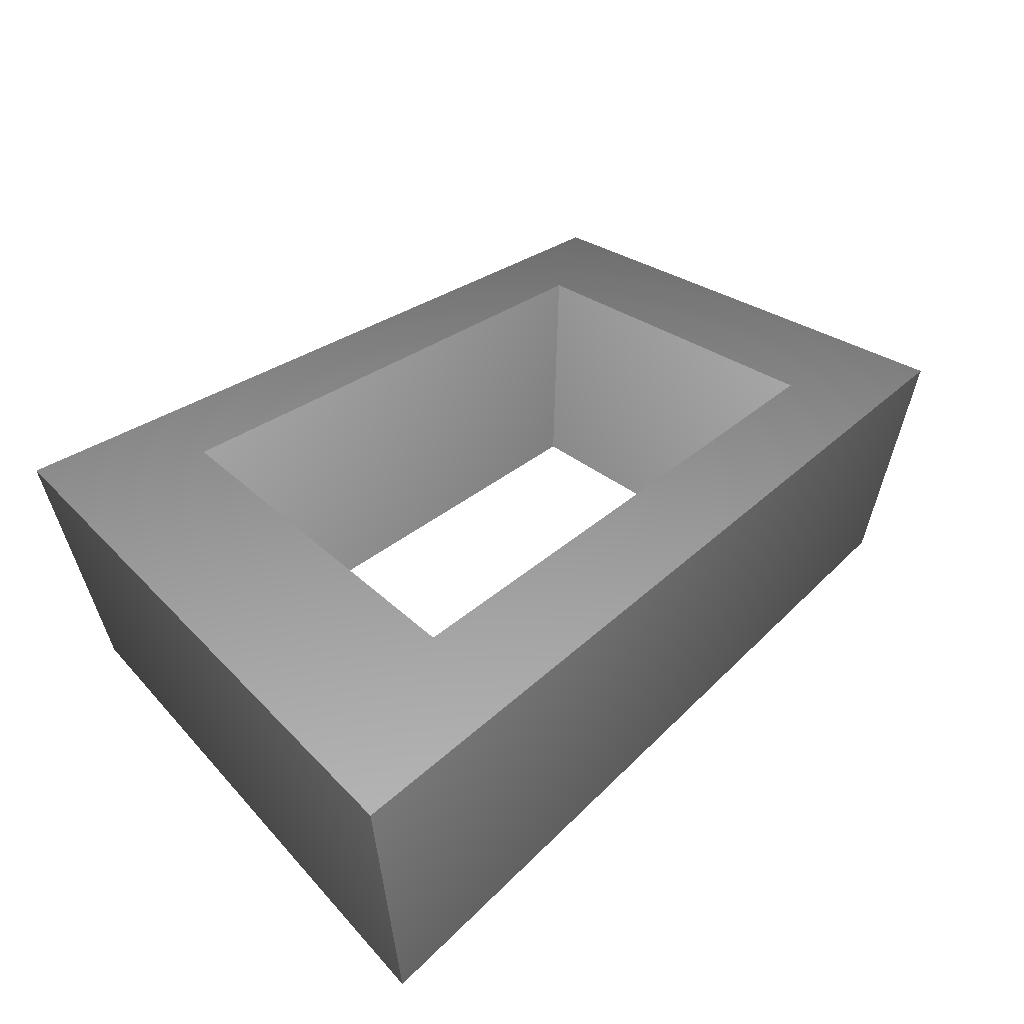}  \\
  & \metrictextsize $\chi=2$   & \metrictextsize \textcolor{correct}{$\chi=2$} & \metrictextsize \textcolor{correct}{$\chi=2$} &  \metrictextsize \textcolor{correct}{$\chi=2$} & \metrictextsize \textcolor{false}{$\chi=0$}  \\
\hline

\begin{minipage}{\promptwidth\textwidth}
\prompttextsize
\vspace*{-1.5 cm}
\input{02-images/00998843/00998843_prompt}
\vspace*{0.0 cm}
\end{minipage}
    &\adjincludegraphics[width=\objectwidth\textwidth, trim={{\trimf\width} {\trimf\width} {\trimf\width} {\trimf\width}},clip]{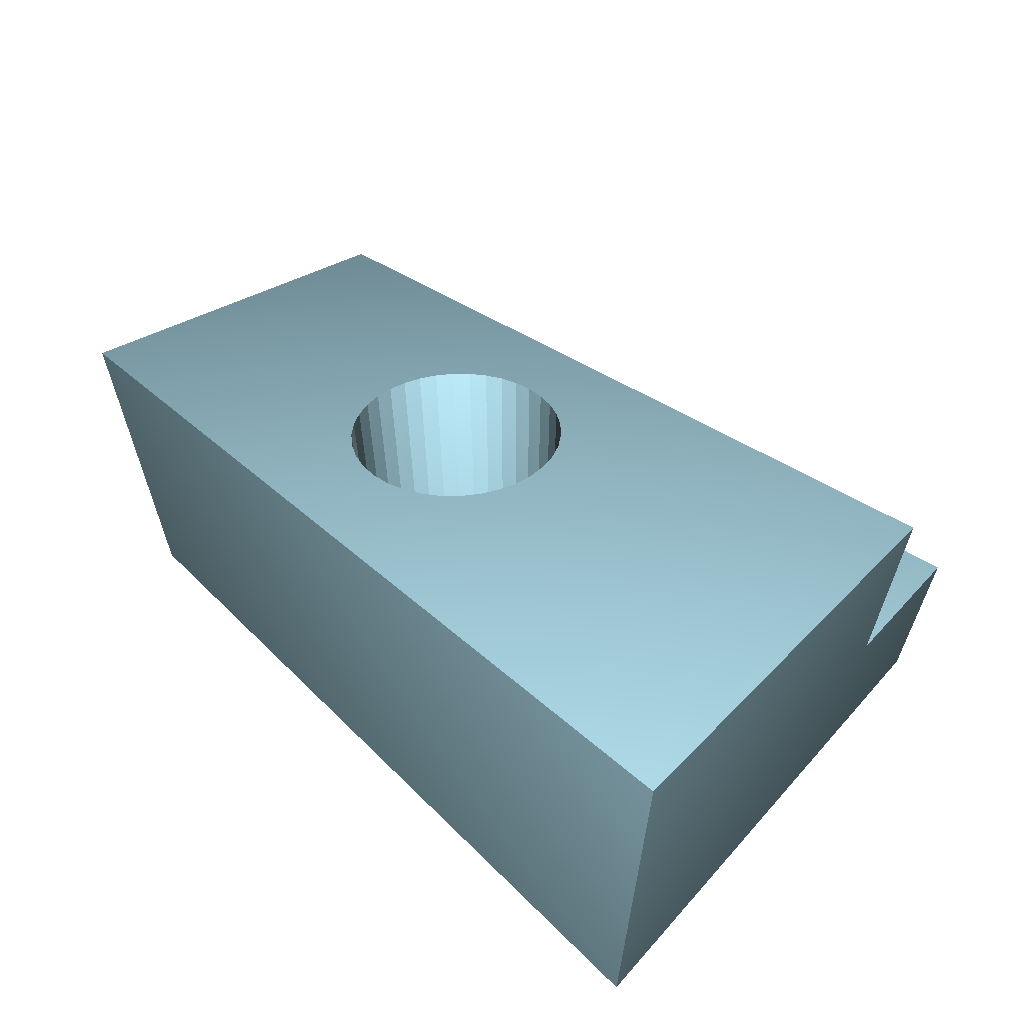}
   & \adjincludegraphics[width=\objectwidth\textwidth, trim={{\trimf\width} {\trimf\width} {\trimf\width} {\trimf\width}},clip]{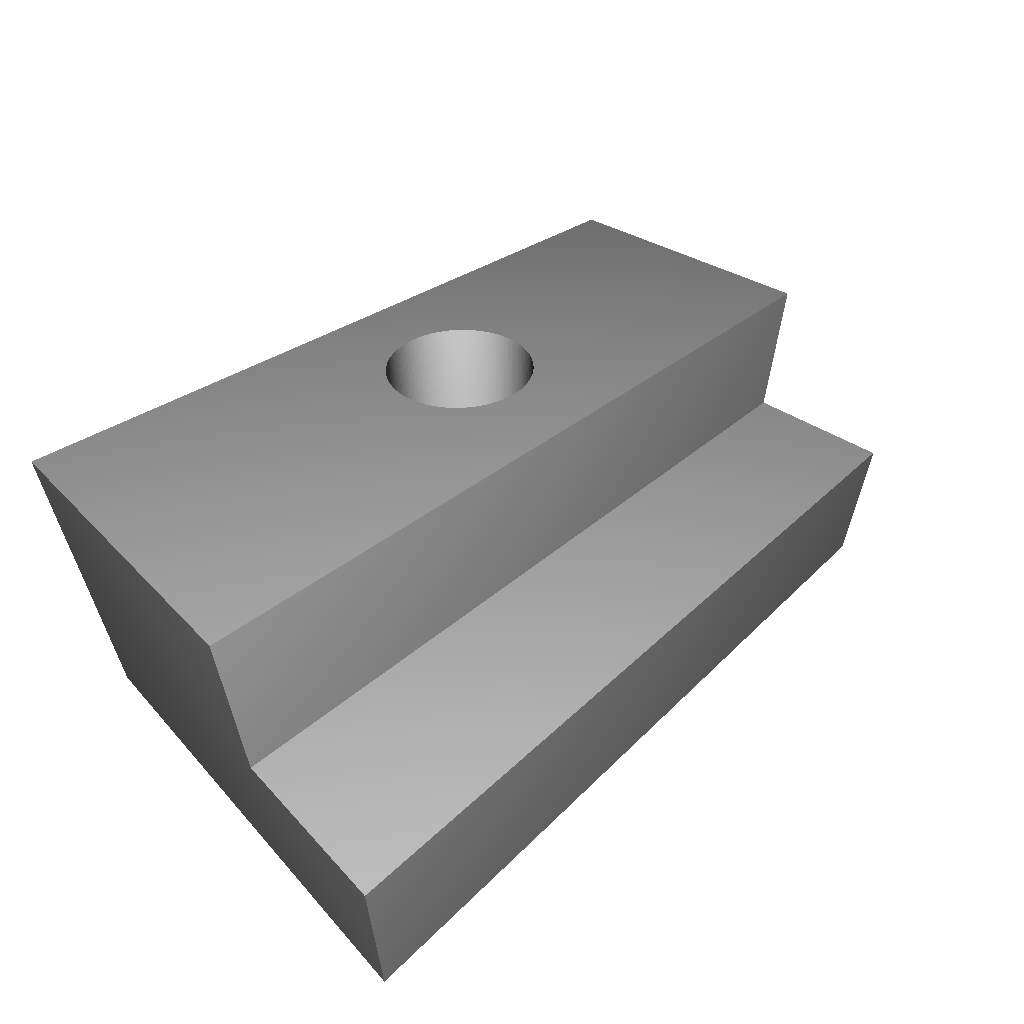} 
 & \adjincludegraphics[width=\objectwidth\textwidth, trim={{\trimf\width} {\trimf\width} {\trimf\width} {\trimf\width}},clip]{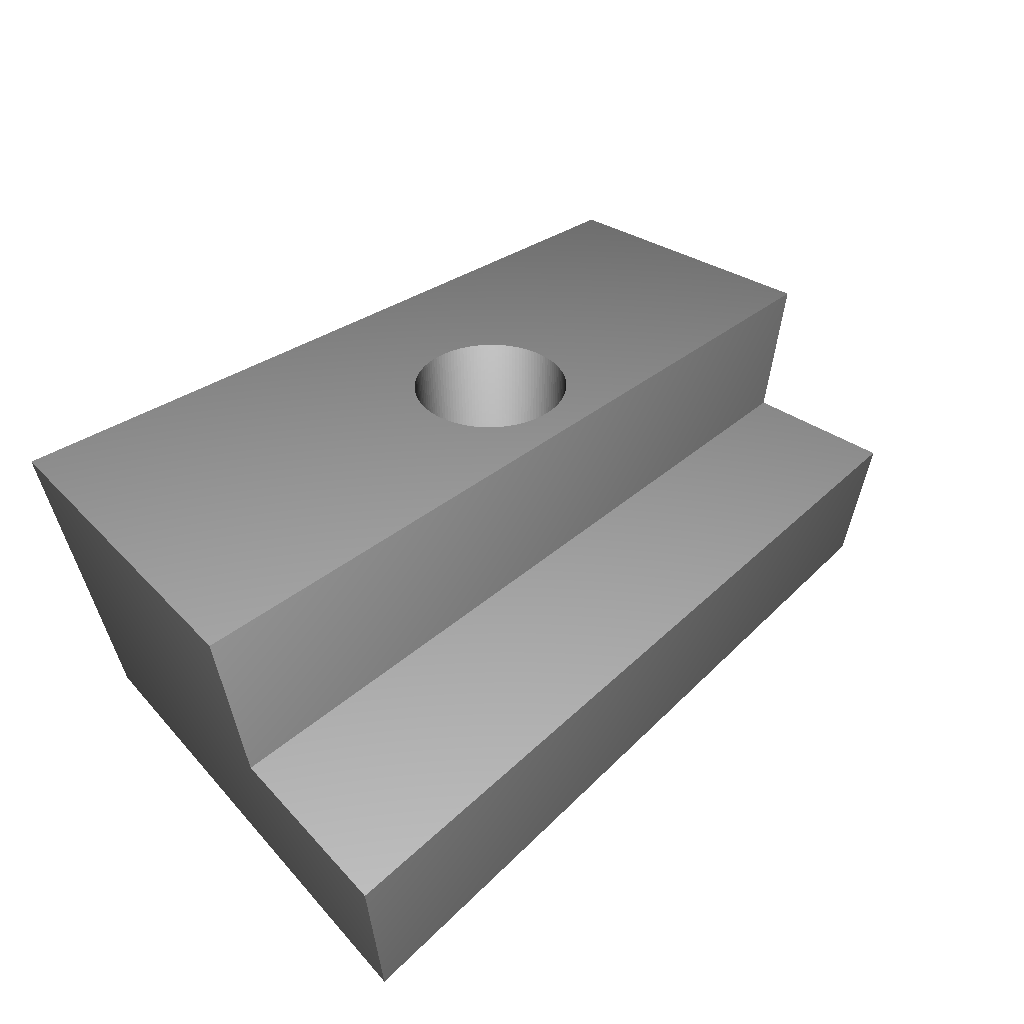} 
 & \adjincludegraphics[width=\objectwidth\textwidth, trim={{\trimf\width} {\trimf\width} {\trimf\width} {\trimf\width}},clip]{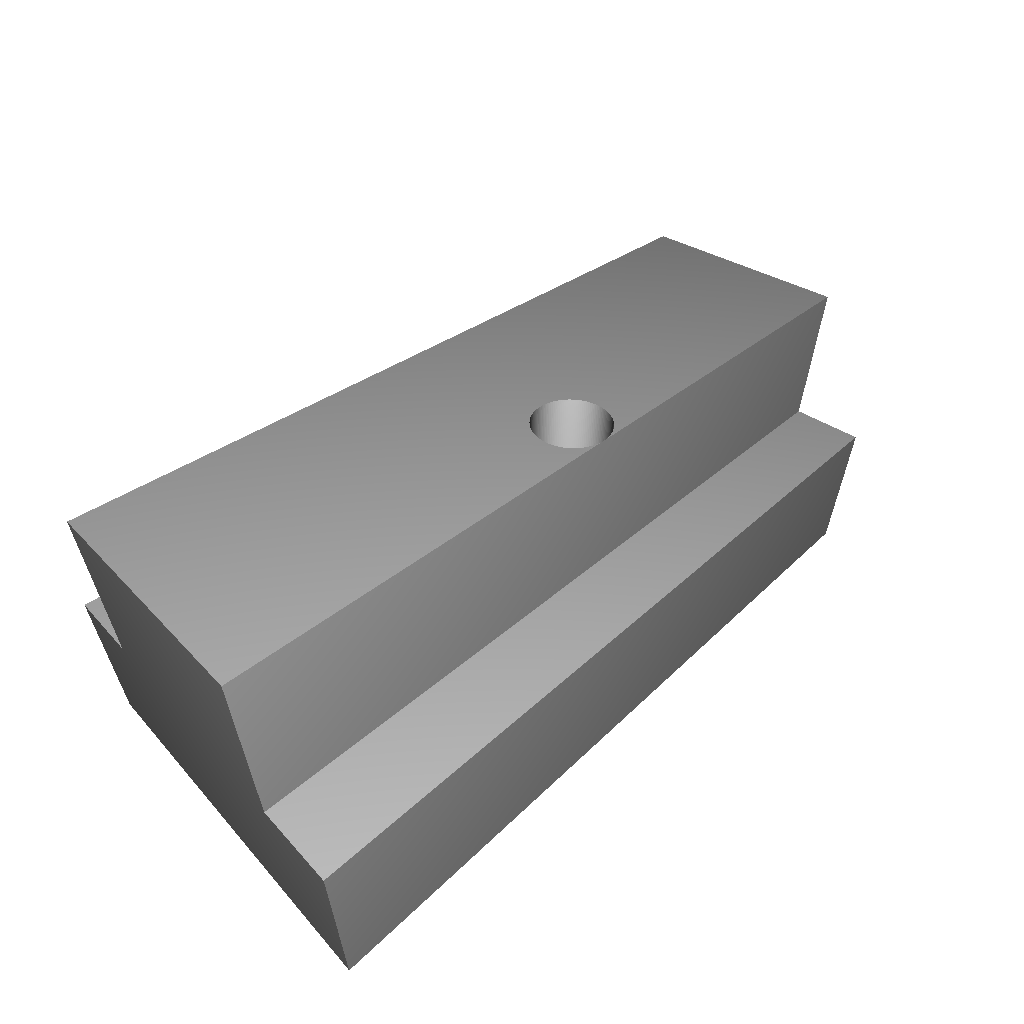} 
 & \adjincludegraphics[width=\objectwidth\textwidth, trim={{\trimf\width} {\trimf\width} {\trimf\width} {\trimf\width}},clip]{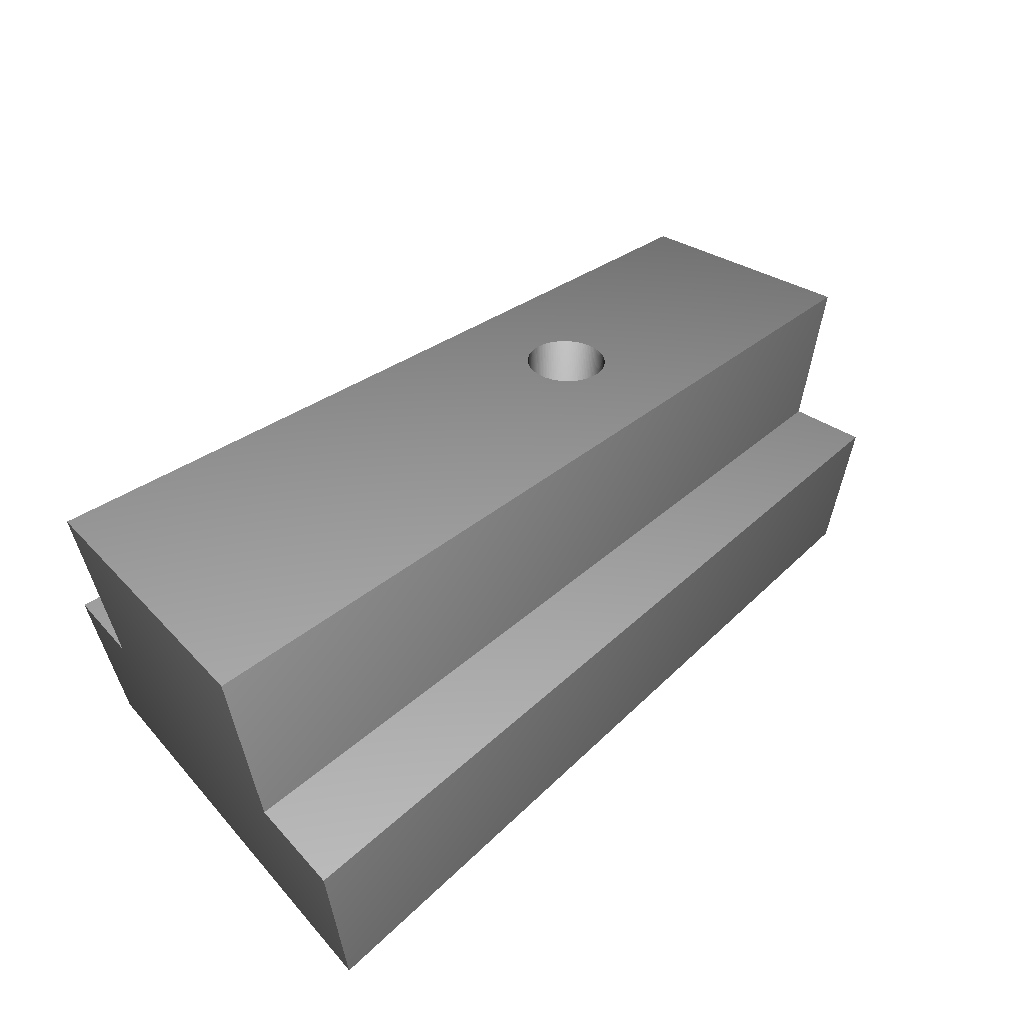} \\
  & \metrictextsize $\chi=0$   & \metrictextsize \textcolor{correct}{$\chi=0$} & \metrictextsize \textcolor{correct}{$\chi=0$} &  \metrictextsize \textcolor{correct}{$\chi=0$} & \metrictextsize \textcolor{correct}{$\chi=0$}  \\
\hline

\begin{minipage}{\promptwidth\textwidth}
\prompttextsize
\vspace*{-1.5 cm}
\input{02-images/00521437/00521437_prompt}
\vspace*{0.0 cm}
\end{minipage}
    &\adjincludegraphics[width=\objectwidth\textwidth, trim={{\trimg\width} {\trimg\width} {\trimg\width} {\trimg\width}},clip]{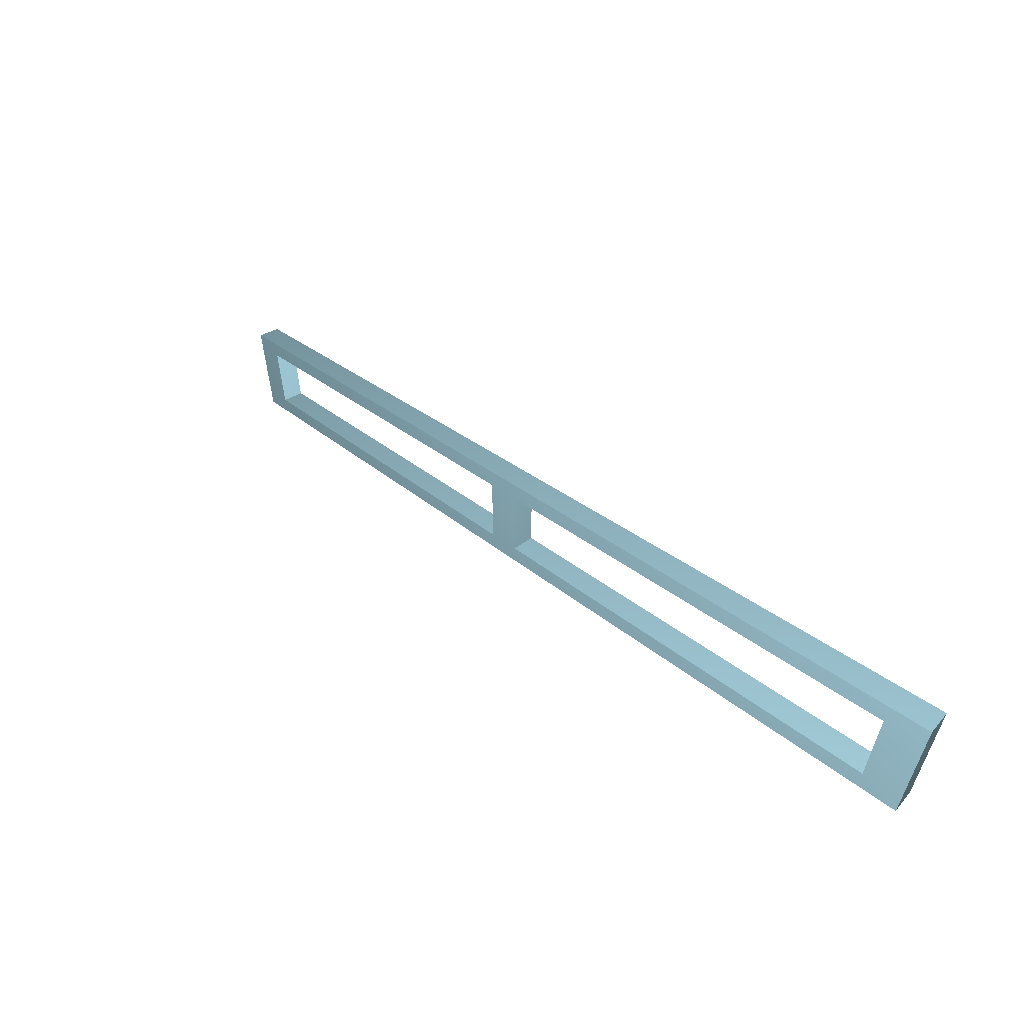}
   & \adjincludegraphics[width=\objectwidth\textwidth, trim={{\trimg\width} {\trimg\width} {\trimg\width} {\trimg\width}},clip]{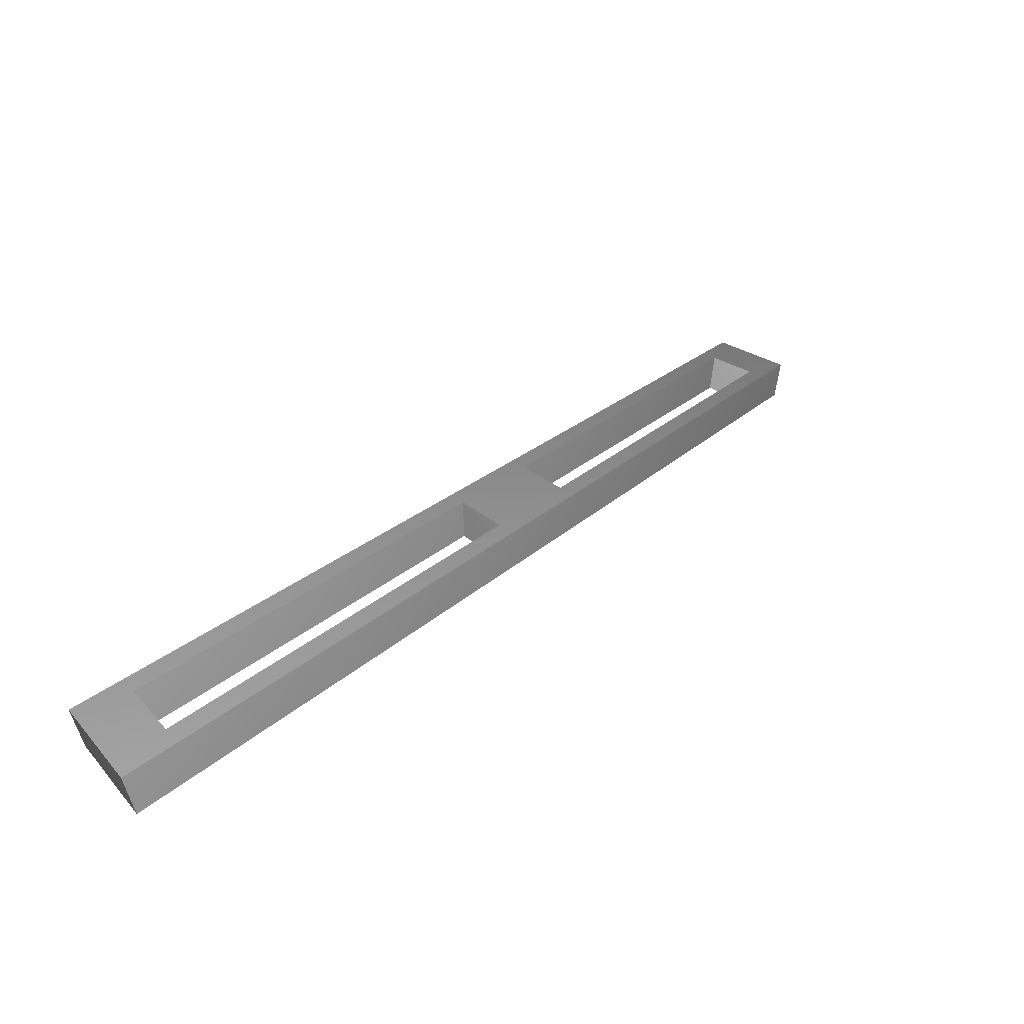}
 & \adjincludegraphics[width=\objectwidth\textwidth, trim={{\trimg\width} {\trimg\width} {\trimg\width} {\trimg\width}},clip]{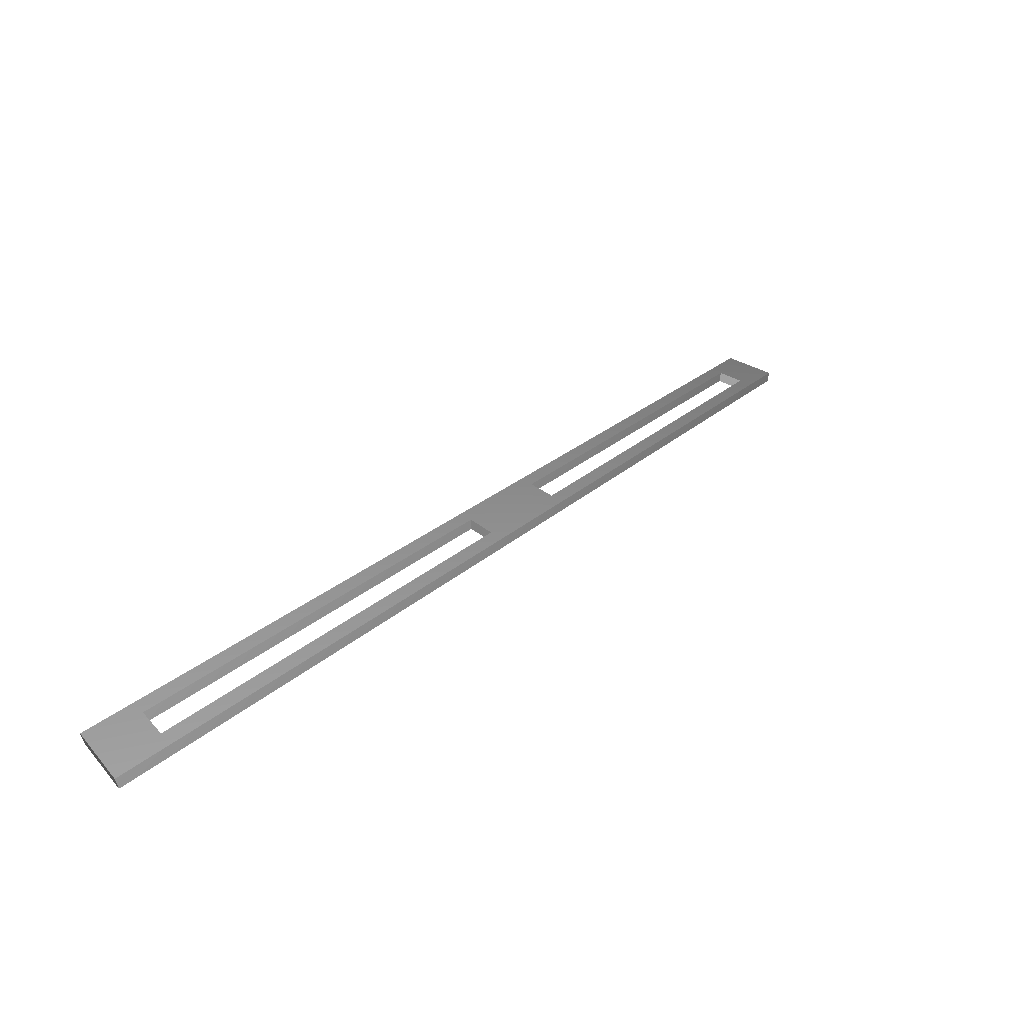} 
 & \adjincludegraphics[width=\objectwidth\textwidth, trim={{\trimg\width} {\trimg\width} {\trimg\width} {\trimg\width}},clip]{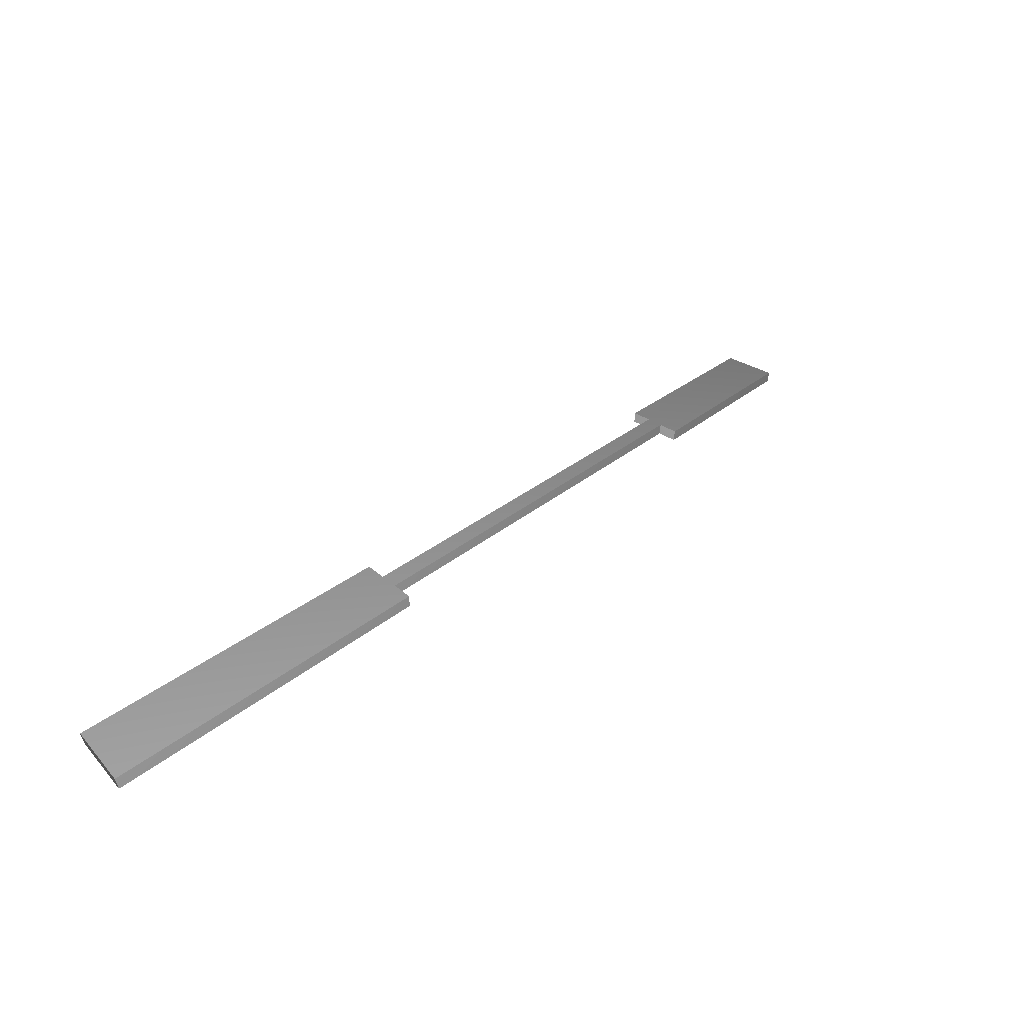} 
 & \adjincludegraphics[width=\objectwidth\textwidth, trim={{\trimg\width} {\trimg\width} {\trimg\width} {\trimg\width}},clip]{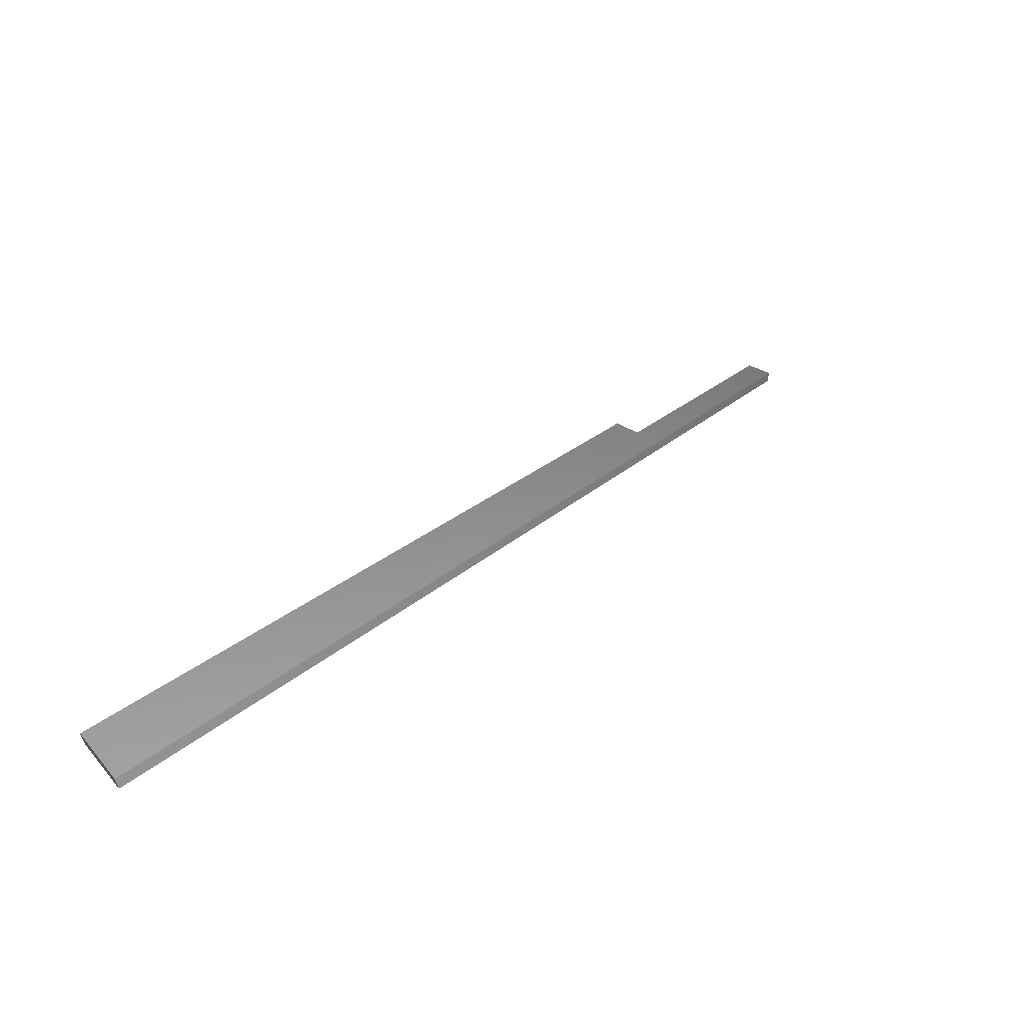} \\
  & \metrictextsize $\chi=-2$  & \metrictextsize \textcolor{correct}{$\chi=-2$} & \metrictextsize \textcolor{correct}{$\chi=-2$} &  \metrictextsize \textcolor{false}{$\chi=2$} & \metrictextsize \textcolor{false}{$\chi=-2$}  \\
\hline

\end{tabular}
}
\caption{Qualitative comparison of our method with prior works. Given a prompt describing a CAD object, our method generates objects that more accurately follow the prompt, align with the ground truth, and exhibit greater topological correctness, as measured by the Euler characteristic $\chi$. 
Green denotes a $\chi$ value that matches the ground truth, while red indicates a deviation.
\label{fig:qualitative}}
\end{figure*}

\textbf{Qualitative evaluation.} 
Fig.~\ref{fig:qualitative} shows the CAD objects generated by our method compared to those generated by other approaches. 
Additionally, the prompt as well as the ground truth object is provided, and below each object, its Euler characteristic $\chi$ is depicted, which we use in our two topology metrics to determine similarity to the ground truth.

Our method demonstrates superior performance specifically in achieving the ground truth $\chi$, which can be intuitively understood as the number of holes in an object.
The first three rows present results where prior methods are missing ``holes''. However, having falsely more ``holes'' than required is also captured by $\chi$ and penalized in our topology metrics, as shown in row four.
For non-watertight objects, neither a closed volume nor a valid $\chi$ exists, making the calculation of volume or topology metrics infeasible.

An example of a non-watertight object is illustrated by the object generated by CADCodeVerify in row four.
In addition to improved topological correctness, our method has the general capability to accurately generate diverse CAD objects that adhere to the geometrical description in the prompt, as demonstrated by the last three rows.




\label{chapter:experiments}

\section{Conclusion}
In this paper, we present EvoCAD, a method for generating CAD objects through CAD code using VLMs and evolutionary optimization.
%
We empirically validated our method using the CADPrompt benchmark dataset and two VLMs, GPT-4V and GPT-4o.
%
Our quantitative and qualitative results demonstrate that our method outperforms previous approaches, such as 3D-Premise and CADCodeVerify, based on multiple metrics, particularly in generating topologically correct objects.

Allowing a quantitative evaluation of topological correctness was also a contribution of this work, as we introduced two novel metrics for comparing 3D objects by utilizing topological properties defined by the Euler characteristic.
This is specifically useful to characterize and compare CAD objects, which often depict complex structures with diverse topological characteristics.  
In this way, our topology-based metrics assess a form of semantic similarity between two objects, thereby complementing existing geometrical metrics that solely evaluate spatial properties.

A current limitation of this work is the relatively small population size and number of optimization generations, because of the growing number of required LLM API calls as these parameters increase. In some samples, this may lead to unrealized potential due to insufficient exploration and early exploitation.
%
%
However, according to Moore's Law, as computational capabilities improve and more efficient LLMs like DeepSeek-R1 \cite{guo2025deepseek} are developed, LLM inference is expected to become progressively cheaper and faster in the future.

In future work, we plan to scale up our evolutionary optimization and evaluate it on increasingly complex data, anticipating even greater advantages over limited-depth refinement methods that plateau after a few steps.


\label{chapter:conclusion}

\FloatBarrier
\printbibliography 

\end{document}